%% file: MA.tex
\documentclass{article}
\usepackage[english]{babel}
\usepackage[a4paper,top=2cm,bottom=2cm,left=3cm,right=3cm,marginparwidth=1.75cm]{geometry}
\RequirePackage{amsthm,amsmath,amsfonts,amssymb}
\usepackage[authoryear,round]{natbib}
\usepackage{graphicx}
\usepackage{enumerate}
\usepackage{url} 
\usepackage{bm}
\usepackage{bbm}
\usepackage{enumitem}
\usepackage{threeparttable}
\usepackage{booktabs}
\usepackage{adjustbox}
\usepackage{algorithm}
\usepackage{algpseudocode}
\usepackage{subcaption}
\usepackage{placeins}
\usepackage{booktabs}
\usepackage{multirow}
\usepackage[colorlinks=true, allcolors=blue]{hyperref}
\newcommand{\Qref}[1]{\hyperlink{question:#1}{\(Q_{#1}\)}}

\newtheorem{assumption}{Assumption}
\newtheorem{condition}{Condition}
\newtheorem{theorem}{Theorem}[section]

\newtheorem{proposition}{Proposition}
\newtheorem{example}{Example}

\newtheorem*{examplecontinued}
{Example~\ref{ex:correlated covariate groups} (continued)}

\title{Non-nested least-squares model averaging: oracle improvement and adaptive candidate model construction}
\date{}
\author{%
	Bosen Cui\textsuperscript{a} and
	Yuhong Yang\textsuperscript{a,b}\\[0.6ex]
	{\small \textsuperscript{a}Yau Mathematical Sciences Center,
		Tsinghua University, Beijing 100084, China}\\
	{\small \textsuperscript{b}Beijing Institute of Mathematical Sciences
		and Applications, Beijing 101408, China}\\[0.6ex]
	{\small Corresponding author: Yuhong Yang
		(\href{mailto:yyangsc@mail.tsinghua.edu.cn}{yyangsc@mail.tsinghua.edu.cn})}%
}

\begin{document}
	\maketitle
	\begin{abstract}
We study when non-nested least-squares model averaging (MA) improves upon model selection (MS) and how candidate construction may preserve these improvements. Given an MA method, a model’s contribution to the average depends on the other models with which it is combined. We introduce residualized model-averaging importance (RMAI) to quantify this dependence and analyze an oracle path that successively admits candidates and reoptimizes the averaging weights. Through an exact decomposition of the oracle gain, together with  conditions on the model’s geometric structure, we characterize three asymptotic regimes: asymptotic equivalence, nonvanishing proportional improvement, and improvement of the risk order. We further establish finite-path approximation for the all-subset averaging oracle under orthogonality and show that, under a correlated design, even small candidate collections can retain a nonvanishing fraction of its gain. These results motivate dynamic model averaging (DMA), a data-driven procedure that adaptively constructs candidates through variable additions, deletions and replacements, evaluating each proposal through joint weight reoptimization. We further derive risk bounds of DMA that account for using the same validation sample to guide the search and estimate the weights. Under orthogonal designs and suitable regularity conditions, DMA achieves a risk bound of \(1+o(1)\) relative to the all-subset averaging oracle based on its fitting sample, while maintaining a nonvanishing proportional improvement over the corresponding MS oracle. Analyses of 32 regression data sets spanning a variety of application scenarios show that, in most cases, DMA outperforms other competing MS methods in terms of predictive performance. This improvement is particularly pronounced in high-dimensional settings.
	\end{abstract}
	\par
\section{Introduction}

Model selection (MS) and model averaging (MA) are two complementary approaches to prediction in the presence of competing model specifications. MS chooses a single candidate model, while MA combines predictions from a set of candidates and leverages their complementary information. A large body of literature studies their statistical performance from both asymptotic and nonasymptotic perspectives. In the asymptotic theory of MS, major lines of work have focused on selection consistency, such as identifying the true model or choosing the better regression procedure \citep{shao1997asymptotic,yang2007consistency}, and on attaining predictive efficiency relative to the best candidate model \citep{shibata1983asymptotic,li1987asymptotic}. Furthermore, for penalized MS,
nonasymptotic oracle inequalities for nonparametric regression based on general linear models have also been established \citep{yang1999model,barron1999risk,birge2001gaussian,baraud2009gaussian};
see \citet{ding2018model} for a related review. For MA, asymptotic optimality has been extensively studied for data-driven combinations on prescribed candidate collections \citep{hansen2007least,hansen2012jackknife,zhang2021new,zhang2023prediction,peng2025optimality,bu2025improving,yu2025unified}. Finite-sample analyses of  aggregation procedures provide risk bounds that quantify the cost of combining candidate estimators \citep{yang2001adaptive,leung2006information,dalalyan2008aggregation,dalalyan2012sharp,wang2014adaptive,bellec2018optimal}. Nonasymptotic oracle inequalities have also been established for Mallows model averaging (MMA), including sharp bounds relative to the best convex combination \citep{peng2025mallows,peng2026optimalrates}. Taken together, these results assess how closely data-driven MA tracks a specified averaging oracle. Understanding the actual benefit of averaging over selection therefore requires a direct comparison between the averaging oracle and the selection oracle, together with an assessment of how the structure of the candidate collection shapes this difference.

\citet{peng2022improvability} studied this oracle comparison for nested models based on orthogonal series expansions,  \citet{xu2025model} extended the analysis to a broader class of nested linear least-squares estimators. Under their respective conditions, nested MA can reduce the selection-oracle risk by a nonvanishing proportion, although the two oracle risks remain of the same asymptotic order. By contrast, non-nested candidate collections allow models based on different predictor subsets to capture complementary components of the regression signal,  suggesting that non-nested MA may offer greater gains over selection than nested MA. However, the extent to which this complementarity translates into risk reduction also depends on the variances and covariances of the estimation errors. A model’s individual risk therefore does not determine its contribution to an average: a model with relatively high individual risk may still improve a combination by capturing signal missed by the other candidates. Understanding these joint effects is central to determining the magnitude and asymptotic order of the gain over selection. 

This dependence also matters for candidate set construction: screening solely by individual predictive accuracy may discard models that would improve a combination. Existing approaches construct candidate model collections through information-criterion screening or marginal-information grouping \citep{yuan2005combining,ando2014model}, LASSO or adaptive LASSO paths \citep{zhang2016optimal,bellec2016aggregation,zhang2020parsimonious}, and sequential screening or residual-based averaging \citep{xie2017heteroscedasticity,lan2018sequential}.  \citet{peng2025optimality} construct reduced nested collections that attain the all-subset benchmark in an orthogonal basis under conditions ensuring asymptotic equivalence between the nested and all-subset averaging oracle risks. \citet{zhu2023scalable}  obtain a scalable candidate representation through singular value decomposition, with all-subset loss guarantees subject to specific regularity conditions on dimensionality and the optimal averaging risk.
These guarantees rely on particular structural conditions or changes in candidate representation. 

When the number of predictors is large, considering all possible subset models is typically infeasible in practice, making it crucially important to study the element of model list construction. In this work, we focus on the joint contributions of non-nested models formed from the original predictors, as a basis for understanding averaging gains and guiding candidate construction. This leads to the following three questions:

\noindent\emph{\hypertarget{question:1}{\(Q_1\)}: When, and by how much, does the averaging oracle improve upon the selection oracle for a non-nested candidate collection?}

\noindent\emph{\hypertarget{question:2}{\(Q_2\)}: How much of the gain available from all-subset averaging can a collection retain under constraints on its size and individual model dimensions?}

\noindent\emph{\hypertarget{question:3}{\(Q_3\)}: Can such a collection be constructed from data while retaining its gains after accounting for the statistical costs of candidate search and weight estimation?}

To address \Qref{1}, we consider possibly growing collections of least-squares models under fixed-design.
Unlike nested construction, non-nested candidates do not admit a single orthogonal decomposition, thus the oracle risk difference between the MA and MS cannot be evaluated directly from individual model risks. We overcome such an obstacle by analyzing an oracle path connects the MS oracle to the MA oracle through sequential candidate admissions, decomposing the total improvement into incremental risk reductions. We introduce  residualized model-averaging importance (RMAI) to quantify a candidate's potential contribution to the average.
We characterize these risk reductions in terms of RMAI and the projection innovation contributed by each admission. Their accumulation distinguishes three asymptotic regimes: asymptotically equivalent oracle risks, a nonvanishing proportional improvement with the same risk order, and an improvement in risk order. Further analysis shows how the decay of ranked RMAI values and the geometry of the candidate models jointly determine these regimes.

To answer \Qref{2}, we investigate how much of the gain from all-subset MA can be retained under constraints on both the number of candidate models and the dimension of each model. Under orthogonal designs, we establish an upper bound on the relative excess risk that decreases inversely with the number of admitted models, independently of the number of predictors. Moreover, since a nested collection can only cover the predictors in its largest model, a small non-nested collection can cover additional weak signals complementarily, asymptotic attaining the all-subset oracle risk while achieving a nonvanishing proportional risk reduction over the best admissible nested collection. Under correlated design, a small non-nested collection likewise can preserve a positive fraction of the risk reduction that all-subset averaging achieves over the MS oracle.

Guided by these oracle insights, we answer \Qref{3} by developing dynamic model averaging (DMA), a data-driven procedure that constructs candidates adaptively through  predictor additions, deletions and replacements. Candidate models are estimated on a fitting sample and then evaluated on a validation sample through the prediction error reduction after joint weight reoptimization. We derive finite sample prediction-risk bounds that account for the repeated use of validation data in both candidate search and weight estimation, linking empirical improvement to risk reduction. Under a Gaussian orthogonal setting with a positive fraction of signals at the fitting-noise scale, DMA's expected prediction risk is at most \(1+o(1)\) times the all-subset oracle risk based on its fitting sample, while retaining a nonvanishing proportional improvement over the corresponding MS oracle. Since  DMA involving sample splitting, these oracle risks are defined on the fitting subsample rather than the full sample.

 To examine whether DMA nevertheless outperforms existing methods that using the full sample, we assess its finite-sample performance of DMA through simulations and analyses of 32 regression data sets. Simulations under polynomial and exponential coefficient decay show predictive gains over the nested Mallows and jackknife averaging procedures considered, with improvements over MS methods in most settings. In the real-data analyses, DMA achieves lower average test mean squared error (MSE) than all six MS competing methods on 26 of the 32 data sets. The improvements are significant in the high-dimensional group with a median reduction in test MSE of 7.50\% relative to LASSO and median reductions ranging from 13.73\% to 20.65\% relative to the other five MS methods. These findings demonstrate the practical predictive benefits of DMA across different signal and dimensional settings.

Section~\ref{sec:setup} introduces the basic setting, including the candidate collections and oracle benchmarks. Section~\ref{sec:rmai} develops RMAI and the oracle reoptimization path. Section~\ref{sec:main-results} characterizes the regimes of oracle improvement and examines how small candidate collections retain averaging gains. Section~\ref{sec:data-driven-path} presents DMA and establishes its prediction-risk bounds and all-subset oracle guarantees. Sections \ref{sec:simulation} and \ref{sec:real-data} report simulation studies and empirical analyses, respectively, and Section \ref{sec:: concolusion} concludes.
The Supplementary Material contains the proofs, further theoretical results and additional simulation details.

\section{Problem set-up}
\label{sec:setup}

Consider the linear regression model
\[
	Y_i=\mu_i+\epsilon_i=\mathbf{X}_i^T\bm\beta+\epsilon_i,
	\qquad i=1,\ldots,n,
\]
where $\mathbf{X}_i\in\mathcal{R}^{p_n}$ is the covariates vector and $\bm\beta\in\mathcal{R}^{p_n}$ denotes the respective coefficient vector. Write 
$
\mathbf{Y}=(Y_1,\ldots,Y_n)^T,$ $
\bm{\mu}=(\mu_1,\ldots,\mu_n)^T,$ and $\bm{\epsilon}=(\epsilon_1,\ldots,\epsilon_n)^T$. Let $
\mathcal{X}=(\mathbf{X}_1,\ldots,\mathbf{X}_n)^T
$ denote the design matrix. Throughout the paper, we treat the design matrix \(\mathcal X\) as fixed and assume that \(E(\bm\epsilon)=\bm0\) and \(E(\bm\epsilon\bm\epsilon^T)=\sigma^2I_n\), with \(0<\sigma^2<\infty\).  We use \(\|\cdot\|\) to denote the Euclidean norm.
Let \(\mathcal M_n=\{S_1,\ldots,S_{M_n}\}\) be a family of candidate models, where \(S_m\subseteq\{1,\ldots,p_n\}\) is nonempty and \(k_m=|S_m|\ge1\), for $1 \le m \le M_n$. Both \(p_n\) and \(M_n\) may increase with \(n\). 
For each candidate \(m\), let \(\mathcal X_{(m)}\) contain the columns of \(\mathcal X\) indexed by \(S_m\), and assume that \(\mathcal X_{(m)}^T\mathcal X_{(m)}\) is nonsingular. The ordinary least squares (OLS) estimator is \(\widehat{\bm\mu}_m=P_m\mathbf Y\), where \(P_m=\mathcal X_{(m)}(\mathcal X_{(m)}^T\mathcal X_{(m)})^{-1}\mathcal X_{(m)}^T\). Its prediction risk is
\[
R_n(m)
=\frac1n E\!\left(\|\widehat{\bm\mu}_m-\bm\mu\|^2\right)
=\frac{\|(I_n-P_m)\bm\mu\|^2}{n}+\frac{\sigma^2k_m}{n}.
\]
The MS oracle minimizes this risk over \(\mathcal M_n\). We assume that it is unique and write \(m_n^*=\operatorname*{arg\,min}_{1\le m\le M_n}R_n(m)\). 
Let $\bm{w}$ be the weight vector in the unit simplex in $\mathcal{R}^{M_n}$:
\[
\mathcal H_n=\left\{\bm w\in\mathcal R^{M_n}:w_m\ge0,\ \sum_{m=1}^{M_n}w_m=1\right\}.
\]
Define the averaging projection matrix as  \(P(\bm w)=\sum_{m=1}^{M_n}w_mP_m\) and the corresponding MA estimator as \(\widehat{\bm\mu}(\bm w)=P(\bm w)\mathbf Y\). The risk of $\widehat{\bm\mu}(\bm w)$ can be decomposed into
\[
R_n(\bm w)
=\frac1n E\!\left(\|\widehat{\bm\mu}(\bm w)-\bm\mu\|^2\right)
=\frac{\|(I_n-P(\bm w))\bm\mu\|^2}{n}
+\frac{\sigma^2}{n}\operatorname{tr}\{P(\bm w)^2\}.
\]
The oracle MA weight is any \(\bm w_n^*\in\operatorname*{arg\,min}_{\bm w\in\mathcal H_n}R_n(\bm w)\).  Since the simplex contains every model-selection vertex, \(R_n(\bm w_n^*)\le R_n(m_n^*)\). We measure the oracle improvement of MA over MS by
\[
\Delta_n=R_n(m_n^*)-R_n(\bm w_n^*).
\]
 For any nonempty index set \(\mathcal A\subseteq\{1,\ldots,M_n\}\), define the simplex of averaging weights supported on $\mathcal{A}$ by \(\mathcal H_n(\mathcal A)=\{\bm w\in\mathcal H_n:w_m=0\text{ for }m\notin\mathcal A\}\). 
We distinguish three asymptotic regimes of oracle improvement of non-nested MA over MS:
\begin{enumerate}
	\item[(i)] \emph{Asymptotic equivalence.}
	If
	\[
	\frac{\Delta_n}{R_n(m_n^*)}\longrightarrow 0,
	\]
	then the MA and MS oracle risks are asymptotically equivalent.
	
	\item[(ii)] \emph{Nonvanishing proportional improvement.}
	If
	\[
	\Delta_n\asymp R_n(m_n^*),
	\qquad
	R_n(\bm w_n^*)\asymp R_n(m_n^*),
	\]
	then the averaging oracle achieves a nonvanishing proportional
	reduction in risk, while the two oracle risks remain of the same
	asymptotic order.
	
	\item[(iii)] \emph{Improvement in risk order.}
	If
	\[
	\frac{R_n(\bm w_n^*)}{R_n(m_n^*)}\longrightarrow 0,
	\]
	then the averaging-oracle risk is of smaller asymptotic order
	than the selection-oracle risk.
\end{enumerate}

\section{Residualized model-averaging importance and the oracle path}
\label{sec:rmai}

To understand theoretically how MA improves on MS, we construct an oracle path from the MS oracle to the MA oracle. Along this path, candidates are added sequentially and the averaging weights are jointly reoptimized after each addition. We examine which candidates reduce risk and how these reductions accumulate. To analyze how an individual candidate can improve the averaging, we first decompose the change in risk when the two are combined.

Let \(\bm w\in\mathcal H_n\) satisfy \(R_n(\bm w)\le R_n(m_n^*)\), and consider
an individual candidate  $m\notin \text{supp}(\bm{w})$. For $0\le \eta \le 1$ write the convex combination \(\bm w_m(\eta)=(1-\eta)\bm w+\eta\bm e_m\), where \(\boldsymbol e_m\) is an \(M_n\)-dimensional vector with its \(m\)th entry equal to 1 and all other entries equal to 0. This combination assigns weight \(\eta\) to candidate \(m\) while shrinking each existing weight by $1-\eta$. Define \(s_{m\mid\bm w}=n^{-1}E[\{\bm\mu-\widehat{\bm\mu}(\bm w)\}^{T}\{\widehat{\bm\mu}_m-\widehat{\bm\mu}(\bm w)\}]\) and \(c_{m\mid\bm w}=n^{-1}E[\|\widehat{\bm\mu}_m-\widehat{\bm\mu}(\bm w)\|^2]\).
 The quantity \(s_{m\mid\bm w}\) measures whether moving toward candidate \(m\) corrects the current prediction error, whereas \(c_{m\mid\bm w}\) measures the magnitude of that change. A direct calculation decomposes the risk difference as
\[
R_n\{\bm w_m(\eta)\}-R_n(\bm w)
=-2s_{m\mid\bm w}\eta+c_{m\mid\bm w}\eta^2.
\]
The candidate \(m\) therefore reduces risk for some $\eta \in (0,1)$ if and only if \(s_{m\mid\bm w}>0\). Since $R_n(\bm{w})\le R_n(m_n^*) \le R_n(m)$, evaluating the identity at \(\eta=1\) gives \(c_{m\mid\bm w}-2s_{m\mid\bm w}=R_n(m)-R_n(\bm w)\ge0\). 
For a candidate with $s_{m\mid \bm{w}}>0$, define its \emph{residualized model-averaging importance} (RMAI) relative to the current average by
\[
\theta_{m\mid\bm w}
=\frac{s_{m\mid\bm w}}{c_{m\mid\bm w}-s_{m\mid\bm w}}
=\frac{c_{m\mid\bm w}-\{R_n(m)-R_n(\bm w)\}}
{c_{m\mid\bm w}+\{R_n(m)-R_n(\bm w)\}}
\in(0,1].
\]
With the existing weight proportions held fixed, the optimal weight on the new candidate
is  \(\eta_{m\mid\bm w}^{*}=\theta_{m\mid\bm w}/(1+\theta_{m\mid\bm w})\in (0,1/2] \). Hence RMAI is the ratio of the optimal weight assigned to the candidate to the weight retained on the current average. A value near zero corresponds to a small optimal candidate weight, whereas a value near one corresponds to nearly equal weights on the candidate and the current average. The risk-difference expression gives the same interpretation: RMAI is close to one when $R_n(m) -R_n(\bm{w})$ is small relative to $c_{m \mid \bm{w}}$. 
The corresponding risk reduction can be expressed as
\[
R_n(\bm w)-R_n\{\bm w_m(\eta_{m\mid\bm w}^{*})\} 
=c_{m\mid\bm w}\left(\frac{\theta_{m\mid\bm w}}{1+\theta_{m\mid\bm w}}\right)^2
=(c_{m\mid\bm w}-s_{m\mid\bm w})\frac{\theta_{m\mid\bm w}^{2}}{1+\theta_{m\mid\bm w}}.
\]
Thus RMAI describes the optimal weight for candidate $m$, while the magnitude of the risk reduction also depends on the scale
 \(c_{m\mid\bm w}-s_{m\mid\bm w}=\{c_{m\mid\bm w}+R_n(m)-R_n(\bm w)\}/2\).
  Section~\ref{sec:rmai-regimes} bounds this scale in terms of the signal and projection geometry and uses these bounds, together with RMAI, to characterize how risk reductions accumulate along the oracle path.

In the preceding analysis, the interpretation assumes that the existing weights are reduced proportionally. However, joint reoptimization after admitting a candidate should allow these proportions to change. We therefore construct the oracle path using gains after full weight reoptimization below. Initialize \(\bm w_{0,n}=\bm e_{m_n^*}\) and \(\mathcal A_{0,n}=\{m_n^*\}\). At step $k$, let $\bm{w}_{k,n}$ and $\mathcal{A}_{k,n}$ denote the averaging weights and active set, respectively, and let \(m_{k,n}\) denote the individual candidate selected for admission. After adding $m_{k,n}$, update the weights and active set  by $\bm{w}_{k+1} \in \arg \min_{\bm{w} \in \mathcal{H}_n (\mathcal{A}_{k,n} \cup \{m_{k,n}\})} R_n(\bm{w})$
and $\mathcal{A}_{k+1} = \text{supp} (\bm{w}_{k+1,n})$. For notational simplicity, we write $s_{k,n} = s_{m_{k,n}\mid \bm{w}_{k,n}}$, $c_{k,n} = c_{m_{k,n}\mid \bm{w}_{k,n}}$ and $\theta_{k,n} = \theta_{m_{k,n}\mid \bm{w}_{k,n}}$.

 For each inactive candidate \(m\notin\mathcal A_{k,n}\), define its reoptimized gain by
\[
\gamma_{k,n}(m)
=
R_n(\boldsymbol w_{k,n})
-
\min_{\boldsymbol w\in\mathcal H_n(\mathcal A_{k,n}\cup\{m\})}
R_n(\boldsymbol w),
\]
and write \(\gamma_{k,n}=\gamma_{k,n}(m_{k,n})\) for the selected candidate. Algorithm~\ref{alg:oracle-path} admits a candidate with the largest positive gain. For each weight optimization, if the minimizing weights are nonunique, choose the minimizer
with the smallest Euclidean norm. Algorithm \ref{alg:oracle-path} defines the oracle path analyzed in Sections \ref{sec:rmai} and \ref{sec:main-results}.

\begin{algorithm}[t]
\caption{Construction of the oracle reoptimization path}
\label{alg:oracle-path}
\begin{algorithmic}[1]
\State \(\mathcal A_{0,n}\gets\{m_n^*\}\),
       \(\bm w_{0,n}\gets\bm e_{m_n^*}\),
       and \(k\gets0\)
\Loop
 \If{\(\mathcal A_{k,n}=\{1,\ldots,M_n\}\)}
\State \textbf{break}
\EndIf
    \State Evaluate the reoptimized gain \(\gamma_{k,n}(m)\) for every
           \(m\notin\mathcal A_{k,n}\)
    \If{\(\displaystyle
          \max_{m\notin\mathcal A_{k,n}}\gamma_{k,n}(m)=0\)}
        \State \textbf{break}
    \EndIf
    \State \(\displaystyle
    m_{k,n}\gets
    \min\operatorname*{arg\,max}_{m\notin\mathcal A_{k,n}}
    \gamma_{k,n}(m)\)
    \State Choose \(\displaystyle
    \bm w_{k+1,n}\in
    \operatorname*{arg\,min}_{
    	\bm w\in\mathcal H_n(\mathcal A_{k,n}\cup\{m_{k,n}\})
    }
    R_n(\bm w)\)
    \State \(\mathcal A_{k+1,n}
    \gets\operatorname{supp}(\bm w_{k+1,n})\)
    \State \(k\gets k+1\)
\EndLoop
\State \(K_n\gets k\)
\end{algorithmic}
\end{algorithm}

An individual candidate has positive reoptimized gain if and only if $s_{m|\bm{w}_{k,n}} >0$. Every selected candidate therefore has a well-defined RMAI. Since the path starts at the MS oracle and
decreases risk at every admission, $R_n (\bm{w}_{k,n}) \le R_n(m_n^*)$ throughout. The combination obtained by
proportionally reducing the existing weights is feasible in the joint optimization, so
\[
\gamma_{k,n}
\ge
\max_{\substack{m\notin\mathcal A_{k,n}\\ s_{m\mid\bm w_{k,n}}>0}}
c_{m\mid\bm w_{k,n}}
\left(\frac{\theta_{m\mid\bm w_{k,n}}}{1+\theta_{m\mid\bm w_{k,n}}}\right)^2
\]
at every accepted step. Thus any available candidate with a given risk reduction under proportional rescaling guarantees at least that much progress by Algorithm~\ref{alg:oracle-path}. This property allows signal and design conditions that ensure useful candidates are available to yield bounds on the number of admissions needed for a prescribed risk reduction.

To relate the selected candidate's RMAI to its reoptimized gain, we account for the adjustment available through the existing candidates. Reallocating their weights changes the average through contrasts \(\sum_{i\in\mathcal A_{k,n}}h_i\widehat{\bm\mu}_i\), with \(\sum_{i\in\mathcal A_{k,n}} h_i=0\). Define
\[
q_{k,n}
=\min_{\substack{\bm h\in\mathcal R^{|\mathcal A_{k,n}|}\\ \sum_{i\in\mathcal A_{k,n}}h_i=0}}
\frac1n E\!\left[
\left\|\widehat{\bm\mu}_{m_{k,n}}-\widehat{\bm\mu}(\bm w_{k,n})
+\sum_{i\in\mathcal A_{k,n}}h_i\widehat{\bm\mu}_i\right\|^2
\right].
\]
This quantity measures the prediction difference remaining after its best adjustment by contrasts among the active estimators. Taking all \(h_i=0\) gives \(q_{k,n}\le c_{k,n}\), and optimizing over unrestricted contrasts provides an upper bound on the gain available through weight reoptimization.

\begin{proposition}[Oracle-path representation]
\label{prop:rmai-representation}
For every fixed \(n\), Algorithm~\ref{alg:oracle-path} terminates after finitely many admissions at a model-averaging oracle over \(\mathcal H_n\):
\[
R_n(\bm w_{K_n,n})=R_n(\bm w_n^*).
\]
At every accepted step, \(0<q_{k,n}\le c_{k,n}\) and
\begin{equation}
c_{k,n}\left(\frac{\theta_{k,n}}{1+\theta_{k,n}}\right)^2
\le\gamma_{k,n}
\le\frac{c_{k,n}}{q_{k,n}}\,c_{k,n}\left(\frac{\theta_{k,n}}{1+\theta_{k,n}}\right)^2.
\label{eq:rmai-episode-bound}
\end{equation}
Moreover,
\[
\Delta_n=R_n(m_n^*)-R_n(\bm w_n^*)=\sum_{k=0}^{K_n-1}\gamma_{k,n}.
\]
\end{proposition}

Proposition \ref{prop:rmai-representation} expresses the total oracle improvement through a sum of risk reductions. At each step, preserving the existing weight proportions provides the lower bound, while \(c_{k,n}/q_{k,n}\) bounds its amplification through full reoptimization. If this ratio is uniformly bounded, direct and reoptimized gains have the same order along the path. RMAI therefore provides a common measure of weighting strength. The same representation describes approximation before termination: after \(t\) admissions,
\[
R_n(\bm w_{t,n})-R_n(\bm w_n^*)
=\sum_{k=t}^{K_n-1}\gamma_{k,n}
\le\sum_{k=t}^{K_n-1}\frac{c_{k,n}^{2}}{q_{k,n}}\left(\frac{\theta_{k,n}}{1+\theta_{k,n}}\right)^2,
\qquad 0\le t\le K_n.
\]
The corresponding lower bound replaces \(c_{k,n}^{2}/q_{k,n}\) with \(c_{k,n}\). These bounds relate approximation error to the contributions of the remaining admissions. Since  the path has visited at most \(t+1\) distinct models after \(t\) admissions, they also quantify the accuracy attainable with a small candidate collection. 
The proof of Proposition~\ref{prop:rmai-representation} and its supporting lemmas are given in the Supplementary Material.

\section{Oracle risk improvement of non-nested MA}
\label{sec:main-results}

\subsection{Oracle improvement along the reoptimization path}
\label{sec:rmai-regimes}

Section~\ref{sec:rmai} expresses the improvement of MA as the sum of gains along the oracle reoptimization path. To address \Qref{1}, we now characterize the three asymptotic regimes introduced in Section \ref{sec:setup} by examining each reduction relative to the current risk. Under the conditions below, we relate these proportional reductions to RMAI and projection geometry and use the resulting bounds to compare the two oracle risks.
 Write \(P_{k,n}=P(\bm w_{k,n})\) and define
\[
\rho_{k,n}
=\frac{k_{m_{k,n}}-\operatorname{tr}(P_{k,n}P_{m_{k,n}})}
{\operatorname{tr}(P_{k,n}^{2})}.
\]
The numerator measures the part of the candidate projection that is not fully represented by the current average, while the denominator determines the variance of that average. Thus \(\rho_{k,n}\) measures the relative scale of the candidate's projection contribution. As averaging reduces the current variance, this relative contribution can increase even when RMAI decreases.
The following assumptions connect these two quantities to proportional risk reduction.

\begin{assumption}[Least-squares scale balance]
\label{ass:least-squares-scale-balance}
There exist constants \(C_B,C_S<\infty\) and \(0\leq\delta<1\) such that, uniformly over \(0\leq k<K_n\),
\[
\|(I_n-P_{k,n})\bm\mu\|^2
\leq C_B\sigma^2\operatorname{tr}(P_{k,n}^{2})
\]
and
\[
-\delta\sigma^2\rho_{k,n}\operatorname{tr}(P_{k,n}^{2})
\leq\bm\mu^TP_{k,n}(I_n-P_{m_{k,n}})\bm\mu
\leq C_S\sigma^2\rho_{k,n}\operatorname{tr}(P_{k,n}^{2}).
\]
\end{assumption}

\begin{assumption}[Stable oracle reoptimization]
	\label{ass:stable-reoptimization}
	There exists \(C_F<\infty\) such that \(c_{k,n}/q_{k,n}\leq C_F\) uniformly over \(0\leq k<K_n\).
\end{assumption}
For Assumption \ref{ass:least-squares-scale-balance},
the first condition bounds the bias of the averaged estimator relative to its variance along the oracle path. The second condition controls the magnitude of signal relative to the projection term: its negative part cannot cancel the projection contribution, and its positive part can increase that contribution by a bounded factor. Assumption \ref{ass:stable-reoptimization} requires at least a fixed proportion of $c_{k,n}$, the mean squared prediction difference between the new candidate and the current average, to remain after joint reoptimization.  The following example shows that both assumptions can hold along the oracle path when every candidate model leaves part of the signal unexplained.

\begin{example} [Correlated covariate groups]\label{ex:correlated covariate groups} 
	Let \(H\) be a set of \(h_n\ge 1\)  covariates and \(G_1,\ldots,G_{L_n}\) be disjoint groups of the remaining covariates. A candidate model consists of $H$ and any subset of a single group $G_i$, including the empty subset, for any $1\le i \le L_n$. Write \(r_{i,n}=|G_i|\) and \(r_n=\max_{1\le i\le L_n}r_{i,n}\). Assume \(L_n,r_n\to\infty\), \(r_{i,n}\asymp r_n\) uniformly in \(i\), and the smallest group is unique. Furthermore, \(h_n+\sum_{i=1}^{L_n}r_{i,n}\le n\).
	The design matrix \(\mathcal X_H\) for \(H\) has full column rank and is orthogonal to the design matrix \(\mathcal X_{i,n}\) for each group \(G_i\). Let \(\boldsymbol\zeta_{i,n}=\mathbf 1_{r_{i,n}}/\sqrt{r_{i,n}}\) and suppose that the group Gram blocks satisfy
	\[
	\frac{1}{n}\mathcal X_{i,n}^{T}\mathcal X_{j,n}
	=
	\begin{cases}
		I_{r_{i,n}}, & i=j,\\
		(1-\tau_n)\boldsymbol\zeta_{i,n}\boldsymbol\zeta_{j,n}^{T}, & i\ne j,
	\end{cases}
	\]
	where \(\tau_n\in(0,1)\) and \(\tau_n\asymp L_n^{-1}\). The groups are therefore correlated only through their mean directions \(\boldsymbol v_{i,n}=\mathcal X_{i,n}\boldsymbol\zeta_{i,n}/\sqrt n\). Let \(\boldsymbol\mu_n=\boldsymbol\mu_{H,n}+a_n\boldsymbol u_n\), where \(\boldsymbol\mu_{H,n}\in\operatorname{col}(\mathcal X_H)\) and \(\boldsymbol u_n\) is the unit vector in the direction of \(\sum_{i=1}^{L_n}\boldsymbol v_{i,n}\). Each full-group model \(H\cup G_i\) leaves the same unexplained signal energy \(B_n=a_n^2\tau_n(1-L_n^{-1})\). Assume
	\[
	1<\liminf_{n\to\infty}\frac{B_n}{\sigma^2r_n}
	\le \limsup_{n\to\infty}\frac{B_n}{\sigma^2r_n}<\infty.
	\]
	For sufficiently large \(n\), the smallest full-group model is the unique MS oracle. Algorithm \ref{alg:oracle-path} admits the remaining full-group models in nondecreasing order of size, retains positive weights on all admitted models, and terminates after \(K_n=L_n-1\) admissions. Uniformly over \(0\le k<K_n\),
	\[
	\frac{\|(I_n-P_{k,n})\boldsymbol\mu_n\|^2}
	{\sigma^2\operatorname{tr}(P_{k,n}^2)}
	\lesssim 1,
	\qquad
	\frac{\boldsymbol\mu_n^{T}P_{k,n}(I_n-P_{m_{k,n}})\boldsymbol\mu_n}
	{\sigma^2\rho_{k,n}\operatorname{tr}(P_{k,n}^2)}
	\asymp 1,
	\qquad
	q_{k,n}=c_{k,n}.
	\]
	Thus Assumptions \ref{ass:least-squares-scale-balance} and \ref{ass:stable-reoptimization} hold along the oracle path. The proof is given in Section~\ref*{app:correlated-groups-example} of the Supplementary Material.
\end{example}

Under these assumptions, the bounds in Proposition~\ref{prop:rmai-representation} yield
\begin{equation}
\frac{\gamma_{k,n}}{R_n(\bm w_{k,n})}
\asymp\rho_{k,n}\theta_{k,n}^{2}.
\label{eq:fractional-gain-comparison}
\end{equation}
Consequently, RMAI and projection contribution jointly determine the proportional improvement at each step. Their cumulative effect characterizes the comparison between the selection and averaging oracles.

\begin{theorem}[Oracle-risk regimes]
\label{thm:oracle-risk-regimes}
Under Assumptions~\ref{ass:least-squares-scale-balance} and \ref{ass:stable-reoptimization}, suppose \(K_n\geq1\) for all sufficiently large \(n\). Then
\[
\log\frac{R_n(m_n^*)}{R_n(\bm w_n^*)}
\asymp\sum_{k=0}^{K_n-1}\rho_{k,n}\theta_{k,n}^{2}.
\]
In particular:
\begin{enumerate}
\item[(i)] The oracle risks are asymptotically equivalent if and only if \(\sum_{k=0}^{K_n-1}\rho_{k,n}\theta_{k,n}^{2}\to0\).
\item[(ii)] Averaging achieves a nonvanishing proportional improvement with the same risk order if and only if \(\sum_{k=0}^{K_n-1}\rho_{k,n}\theta_{k,n}^{2}\asymp1\).
\item[(iii)] Averaging improves the risk order if and only if \(\sum_{k=0}^{K_n-1}\rho_{k,n}\theta_{k,n}^{2}\to\infty\).
\end{enumerate}
\end{theorem}

Theorem \ref{thm:oracle-risk-regimes} shows that candidates with small RMAI values can collectively improve the risk order; the key question is whether their corresponding contributions $\rho_{k,n}\theta_{k,n}^{2}$ can accumulate sufficiently. When their sum remains of constant order, MA achieves a nonvanishing proportional improvement. When the sum diverges, the MA oracle risk is of smaller order than the MS oracle risk. We now return to Example \ref{ex:correlated covariate groups} to illustrate the three oracle-risk regimes characterized in Theorem \ref{thm:oracle-risk-regimes}.
\begin{examplecontinued}
	The common covariates $H$ contribute \(\sigma^2h_n/n\) to both oracle risks. The contributions from the optional groups satisfy
	\[
	R_n(m_n^*)-\frac{\sigma^2h_n}{n}
	\asymp \frac{\sigma^2r_n}{n},
	\qquad
	R_n(\boldsymbol w_n^*)-\frac{\sigma^2h_n}{n}
	\asymp \frac{\sigma^2}{n}\left(1+\frac{r_n}{L_n}\right).
	\]
	Averaging therefore yields a gain of order \(\sigma^2r_n/n\). When \(r_n=o(h_n)\), the common variance dominates and the two oracle risks are asymptotically equivalent. When \(h_n\asymp r_n\), the gain is a nonvanishing fraction of the selection risk, while both risks remain of the same order. When \(h_n=o(r_n)\), averaging improves the risk order.
	Along the oracle path, \(\theta_{k,n}\asymp(k+1)^{-1}\) in all three regimes. The common dimension \(h_n\) enters through the paired projection contributions \(\rho_{k,n}\), giving
	\[
	\sum_{k=0}^{K_n-1}\rho_{k,n}\theta_{k,n}^2
	\asymp
	\log\!\left(1+\frac{r_n}{h_n+1+r_n/L_n}\right).
	\]
	This quantity tends to zero, remains of constant order, or diverges in the three cases above, respectively. Details of the derivation are given in Section~\ref*{app:correlated-groups-example} of the Supplementary Material.
\end{examplecontinued}

To see whether the gain comes mainly from a few admissions with large RMAI values or builds across many with smaller values, we rank the completed admissions by decreasing RMAI, writing \(\theta_{[1],n}\geq\cdots\geq\theta_{[K_n],n}\), and keep each \(\rho_{[j],n}\) paired with its corresponding admission. Examining the decay of \(\theta_{[j],n}\) together with the behavior of the paired \(\rho_{[j],n}\) allows us to quantify how the oracle risk ratio varies with \(K_n\). The next result gives explicit rates under polynomial and exponential RMAI decay and establishes lower bounds on the improvement under weaker assumptions.

\begin{theorem}[Oracle improvement under polynomial and exponential RMAI profiles]
\label{thm:fast-rmai-regimes}
Under the assumptions of Theorem \ref{thm:oracle-risk-regimes}, suppose that \(K_n\to\infty\). All profile comparisons below hold uniformly over \(1\leq j\leq K_n\), with constants independent of \(j\) and \(n\). The conditions on \(\rho_{[j],n}\) allow these paired projection contributions to grow with \(j\).
\begin{enumerate}
\item[(i)] \emph{Polynomial RMAI decay.} Suppose that, for some \(\alpha>0\) and \(\xi\in\mathcal R\),
\[
\theta_{[j],n}\asymp\theta_{[1],n}j^{-\alpha},
\qquad
\rho_{[j],n}\asymp\rho_{[1],n}j^\xi.
\]
Then
\[
\log\frac{R_n(m_n^*)}{R_n(\bm w_n^*)}
\asymp\rho_{[1],n}\theta_{[1],n}^2
\begin{cases}
1,&2\alpha>\xi+1,\\
\log K_n,&2\alpha=\xi+1,\\
K_n^{\xi+1-2\alpha},&2\alpha<\xi+1.
\end{cases}
\]
The expression on the right tends to zero if and only if the two oracle risks are asymptotically equivalent, is of constant order if and only if averaging gives a same-order nonvanishing improvement, and diverges if and only if averaging improves the risk order.

\item[(ii)] \emph{Exponential RMAI decay.} Suppose that, for some \(\alpha>0\), \(\xi\in\mathcal R\), and \(C_\rho<\infty\),
\[
\theta_{[j],n}\asymp\theta_{[1],n}e^{-\alpha(j-1)},
\qquad
\rho_{[j],n}\leq C_\rho\rho_{[1],n}j^\xi.
\]
Then
\[
\log\frac{R_n(m_n^*)}{R_n(\bm w_n^*)}
\asymp\rho_{[1],n}\theta_{[1],n}^2.
\]
The two oracle risks are asymptotically equivalent if and only if \(\rho_{[1],n}\theta_{[1],n}^2\to0\), and averaging gives a same-order nonvanishing improvement if and only if \(\rho_{[1],n}\theta_{[1],n}^2\asymp1\). Moreover, \(\liminf_{n\to\infty}R_n(\bm w_n^*)/R_n(m_n^*)>0\).
\end{enumerate}
For part (i) with \(2\alpha\leq\xi+1\), retain the polynomial RMAI profile and suppose only that \(\rho_{[j],n}\geq c_\rho\rho_{[1],n}j^\xi\) for a constant \(c_\rho>0\), \(\xi>0\), and \(\liminf_n\rho_{[1],n}>0\). The bias--variance bound in Assumption \ref{ass:least-squares-scale-balance} alone then implies, for some constant \(c>0\) and all sufficiently large \(n\),
\[
\log\frac{R_n(m_n^*)}{R_n(\bm w_n^*)}
\geq c\rho_{[1],n}\theta_{[1],n}^2
\begin{cases}
\log K_n,&2\alpha=\xi+1,\\
K_n^{\xi+1-2\alpha},&2\alpha<\xi+1.
\end{cases}
\]
\end{theorem}

Theorem \ref{thm:fast-rmai-regimes} shows that oracle improvement depends on the accumulation of \(\rho_{[j],n}\theta_{[j],n}^{2}\). RMAI measures the relative potential of each admission, while \(\rho_{[j],n}\) determines its contribution on the current risk scale. Under exponential RMAI decay and the stated bound on the paired projection contributions, the total contribution remains comparable to \(\rho_{[1],n}\theta_{[1],n}^{2}\), yielding asymptotic equivalence when this quantity vanishes and a nonvanishing proportional improvement when it remains of constant order.
Under polynomial RMAI decay, the accumulation of gains depends on the paired projection contributions and the leading scale \(\rho_{[1],n}\theta_{[1],n}^{2}\). When \(\rho_{[1],n}\theta_{[1],n}^{2}\asymp1\), the critical case \(2\alpha=\xi+1\) gives \(R_n(\bm w_n^*)/R_n(m_n^*)\leq K_n^{-c}\) for some constant \(c>0\). Thus the oracle risk ratio is bounded by a negative power of the path length \(K_n\). When \(2\alpha<\xi+1\), the bound strengthens to \(R_n(\bm w_n^*)/R_n(m_n^*)\leq\exp\{-cK_n^{\xi+1-2\alpha}\}\). This upper bound is exponential in \(K_n^{\xi+1-2\alpha}\). Both cases also allow risk-order improvement when the leading contribution vanishes, provided the cumulative contribution diverges.

The weaker conditions in the theorem preserve these risk upper bounds because each reoptimized admission achieves at least the improvement available from its optimal mixture with the current average. The path therefore connects local opportunities for averaging to global oracle gains. Proofs and a further extension for rapidly decaying ordered RMAI values are given in the Supplementary Material.

\subsection{Oracle gains from small candidate collections}
\label{sec:nonnested-comparison}

Section~\ref{sec:rmai-regimes} characterizes the gains available within a candidate family. To address \Qref{2}, we now ask whether a small collection can retain the gain from all-subset averaging, and which candidate configurations make this possible. Specialize the family in Section~\ref{sec:setup} to all nonempty subsets of \(\{1,\ldots,p_n\}\), retaining the definitions of \(m_n^*\), \(\bm w_n^*\), and \(\Delta_n\). For \(d_n\in\{1,\ldots,p_n\}\), let \(\mathcal I_n(d_n)=\{m:1\le m\le 2^{p_n}-1,\ k_m\le d_n\}\) be the index set of models containing at most \(d_n\) predictors. Apply Algorithm~\ref{alg:oracle-path} to this set, initialized at \(m_n^*\), whenever \(k_{m_n^*}\le d_n\). Its output after at most \(t\) admissions, denoted by \(\bm w_{t,n}\), combines at most \(t+1\) models. The two budgets therefore constrain the number of candidates and the dimension of each fit.

Nested least-squares averaging cannot improve the risk order of selection within the same collection. \citet[Theorem~1]{xu2025model} establish a one-half lower bound under their structural assumptions. The following proposition shows that this restriction follows from nested projection geometry alone.

\begin{proposition}[A sharp risk bound for nested averaging]
\label{prop:nested-half-risk}
Let \(\mathcal C\) be a nonempty collection of least-squares models whose fitted subspaces are nested and fixed conditional on \(\mathcal X\). Suppose that \(E(\bm\epsilon\mid\mathcal X)=\bm0\) and \(E(\|\bm\epsilon\|^2\mid\mathcal X)<\infty\). Then, for every \(n\),
\begin{equation}
\frac12\min_{m\in\mathcal C}R_n(m)
\le\min_{\bm w\in\mathcal H_n(\mathcal C)}R_n(\bm w)
\le\min_{m\in\mathcal C}R_n(m).
\label{eq:nested-half-risk}
\end{equation}
The constant \(1/2\) is sharp, even for two nonempty nested models with homoscedastic errors.
\end{proposition}

The bound requires neither an ordering of coefficient magnitudes nor a restriction on the shape of the model-risk sequence. Thus nested averaging can reduce the selection-oracle risk by at most one half, regardless of the number of candidates. For collections containing \(m_n^*\), this limitation applies relative to the same all-subset selection benchmark.
 We first study how quickly a small collection can approximate the all-subset averaging oracle.

\begin{assumption}[Orthogonal design and model capacity]
\label{ass:orthogonal-size}
Suppose that \(\mathcal X^T\mathcal X=nI_{p_n}\), \(S_{m_n^*}=\{j:\beta_j^2>\sigma^2/n\}\), \(1\le k_{m_n^*}\le d_n\), and \(1\le\sum_{j=1}^{p_n}\beta_j^2/(\beta_j^2+\sigma^2/n)\le d_n\).
\end{assumption}

The last condition makes the optimal weights representable by an average of nonempty models of dimension at most \(d_n\). It allows the number of nonzero coefficients to exceed \(d_n\).

\begin{theorem}[Approximation by a small collection]
\label{thm:nonnested-comparison}
Under Assumption~\ref{ass:orthogonal-size}, for every integer \(t\ge0\),
\begin{equation}
1\le\frac{R_n(\bm w_{t,n})}{R_n(\bm w_n^*)}
\le1+\frac1{t+1}.
\label{eq:nonnested-finite-path}
\end{equation}
\end{theorem}

Consequently, at most \(\lceil\varepsilon^{-1}\rceil\) candidates suffice to attain risk within a factor \(1+\varepsilon\) of the all-subset oracle, independently of \(p_n\). Retaining its entire improvement over selection to first order requires the separate scale comparison \(R_n(\bm w_n^*)/(t+1)=o(\Delta_n)\). The next example shows how a fixed number of complementary candidates can achieve both oracle-risk equivalence and a strict improvement over nested averaging.

\begin{example}[Complementary coverage]
\label{ex:complementary-coverage}
Suppose that \(\mathcal X^T\mathcal X=nI_{p_n}\). Let \(L\ge2\) be fixed and \(0<a<1/(L-1)\). Consider disjoint sets \(S_{0,n}\) and \(W_n\), with \(|S_{0,n}|=h\) and \(|W_n|=Lr\), such that \(\min_{j\in S_{0,n}}n\beta_j^2/\sigma^2\to\infty\), \(\beta_j^2=a\sigma^2/n\) for \(j\in W_n\), and all remaining coefficients are zero. Here \(h,r\ge1\) may depend on \(n\), and \(h/r\to c_h\in[0,\infty)\). Set \(d_n=h+r\). Then \(S_{m_n^*}=S_{0,n}\) eventually.

Each of the first \(L\) admissions selects a model containing \(S_{0,n}\) and \(r\) previously uncovered weak coordinates. After these admissions, all \(Lr\) weak coordinates receive their optimal inclusion weight \(a/(1+a)\). For comparison, let \(\mathcal C_{\mathrm N}^*(t,d_n)\) minimize the averaging-oracle risk over collections with nested supports that contain \(m_n^*\) and have at most \(t+1\) members, each with at most \(d_n\) predictors. Then
\begin{gather*}
\lim_{n\to\infty}\frac{R_n(\bm w_{L,n})}{R_n(\bm w_n^*)}=1, \quad
\lim_{n\to\infty}
\frac{R_n(\bm w_{L,n})}
{\displaystyle\min_{\bm w\in\mathcal H_n(\mathcal C_{\mathrm N}^*(L,d_n))}R_n(\bm w)}
=\frac{(1+a)c_h+La}{(1+a)c_h+La+(L-1)a^2}<1.
\end{gather*}
\end{example}

The dimension bound limits each fit, but does not limit the union of non-nested supports. A nested collection can cover at most \(r\) weak coordinates beyond \(S_{0,n}\), whereas the complementary collection covers all \(Lr\). Averaging retains these additional coordinates only partially, controlling the variance cost of their inclusion. Although each is too weak to enter the selection oracle, their combined contribution yields a persistent gain.
We next extend this mechanism to correlated designs. The following condition controls conditional projections uniformly over subsets.

\begin{assumption}[Correlated design]
\label{ass:correlated-design}
The design has full column rank, and its inverse normalized Gram matrix \(Q=(\mathcal X^T\mathcal X/n)^{-1}\) satisfies \(Q_{jj}-\sum_{\ell\ne j}|Q_{j\ell}|\ge\delta_Q\) and \(Q_{jj}\le L_Q\) for every \(j\), where \(\delta_Q,L_Q>0\) do not depend on \(n\).
\end{assumption}

Inverse-Gram strict diagonal dominance appears in \citet{meinshausen2007discussion}, and its equivalence to the positive cone condition is established by \citet[Theorem~3]{duan2012lars}. Here the dominance margin and diagonal entries are controlled uniformly in \(n\). The condition permits nonvanishing correlations but is stronger than bounded eigenvalues alone.
For a specified nonempty model \(m_n^\circ\), write \(s_n=\|\bm\beta_{S_{m_n^\circ}^c}\|_0\ge1\), and \(a_n=\|(I_n-P_{m_n^\circ})\bm\mu\|^2/(\sigma^2s_n)\). Thus \(a_n\) measures the average residual signal energy per omitted nonzero coefficient relative to the noise variance.

\begin{theorem}[Retention of gains under correlated designs]
\label{thm:correlated-improvement}
Suppose that Assumption~\ref{ass:correlated-design} holds and \(s_n\|\bm\beta_{S_{m_n^\circ}^c}\|_\infty^2\le B\|\bm\beta_{S_{m_n^\circ}^c}\|^2\) for a fixed \(B>0\). Let \(d_n\le p_n\) satisfy \(d_n-k_{m_n^\circ}\asymp d_n\asymp s_na_n\). There exist constants \(a_0>0\) and \(C_0<\infty\), depending only on the fixed constants in these conditions, such that, if \(a_n\le a_0\), \(s_na_n^2\to\infty\), and \(a_n\min_{j\in S_{m_n^\circ}}n\beta_j^2/\sigma^2\to\infty\), then \(m_n^\circ=m_n^*\) uniquely for all sufficiently large \(n\). For every integer \(t\ge1\) satisfying \((t+1)a_n\ge C_0\),
\begin{align}
R_n(m_n^*)-R_n(\bm w_{t,n})
&\asymp\Delta_n\asymp\frac{\sigma^2}{n}s_na_n^2,
\label{eq:correlated-retained-gain}\\
1-\frac{R_n(\bm w_{t,n})}{R_n(m_n^*)}
&\asymp a_n.
\label{eq:correlated-selection-ratio}
\end{align}
\end{theorem}

The path retains a fixed positive fraction of the all-subset oracle gain. The spread condition distributes residual signal across many coordinates, while the dimension condition leaves room for complementary extensions. The strong-signal condition makes further shrinkage of the initial coordinates negligible at the gain scale. When \(a_n\) stays bounded away from zero, a sufficiently large fixed collection yields a nonvanishing proportional improvement. The Supplementary Material also establishes an improvement of order \((\sigma^2/n)s_na_n^2\) over the best nested collection under the same budgets.

\begin{example}[Coefficient decay]
\label{ex:correlated-decay}
Let \(|\beta|_{(j)}\) denote the \(j\)th largest absolute coefficient. The decay conditions below are imposed directly on these ordered coefficient magnitudes. Coefficient signs and locations are unrestricted.

For exponential decay of the ordered coefficients, suppose that Assumption~\ref{ass:correlated-design} holds and \(|\beta|_{(j)}=ce^{-\lambda j}\), with fixed \(c,\lambda>0\). Define \(j_n=\log(nc^2/\sigma^2)/(2\lambda)\) and assume \(p_n/j_n\to\infty\). Then
\begin{gather*}
R_n(\bm w_n^*)\sim R_n(m_n^*)\sim\frac{\sigma^2j_n}{n}, \qquad
\lim_{n\to\infty}
\frac{R_n(\bm w_{t,n})}
{\displaystyle\min_{\bm w\in\mathcal H_n(\mathcal C_{\mathrm N}^*(t,d_n))}R_n(\bm w)}=1
\end{gather*}
for any \(t\ge0\) and \(k_{m_n^*}\le d_n\le p_n\).

For polynomial decay of the ordered coefficients, suppose that \(|\beta|_{(j)}=cj^{-\alpha}\), with fixed \(c>0\) and \(\alpha>1/2\), and define \(j_n=(nc^2/\sigma^2)^{1/(2\alpha)}\). Assume \(p_n/j_n\to\infty\) and \(d_n/j_n\to\kappa\), where \(\int_0^\infty(1+x^{2\alpha})^{-1}\,dx<\kappa<\infty\). There exist a sufficiently small constant \(\varepsilon_0>0\) and an integer \(M_0\ge2\), depending only on \(\alpha\) and \(\kappa\), such that, if \(\|Q-I_{p_n}\|_\infty\le\varepsilon_0\) and \(t+1\ge M_0\), then \(k_{m_n^*}\le d_n\) eventually and
\begin{equation*}
\limsup_{n\to\infty}
\frac{R_n(\bm w_{t,n})}
{\displaystyle\min_{\bm w\in\mathcal H_n(\mathcal C_{\mathrm N}^*(t,d_n))}R_n(\bm w)}<1.
\end{equation*}
\end{example}

The distinction lies in the amount of signal near the noise scale. Exponential coefficient decay leaves only a bounded number of coefficients whose squared magnitudes are comparable to \(\sigma^2/n\), so partial inclusion cannot alter the leading risk. Polynomial coefficient decay places a number proportional to \(j_n\) in this range, making shrinkage useful across many coordinates. With \(d_n/j_n\) bounded, a nested collection cannot cover all useful directions, whereas complementary candidates can extend coverage without increasing individual model dimensions. Both risks remain of order \(\sigma^2j_n/n\); the improvement is in the leading constant.

\section{Dynamic model averaging}
\label{sec:data-driven-path}

Building on the oracle analysis in Section \ref{sec:main-results}, we address \Qref{3} by introducing dynamic model averaging (DMA), which constructs the candidate collection adaptively from data. Starting from the OLS fit on a selected support, DMA generates candidates by adding, deleting or replacing regressors in models with positive current weights. Each proposal is evaluated by the reduction in the validation criterion after jointly reoptimizing the averaging weights over the enlarged collection. The candidate yielding the largest positive reduction is admitted. All admitted models remain available for subsequent weight optimization, including those whose current weights are zero.
We then establish risk bounds for DMA that account for the repeated use of validation data in candidate construction and weight estimation.

\subsection{Dynamic candidate construction and weight reoptimization}
\label{sec:ddp-construction}

We use separate samples to fit candidate models and evaluate their contributions to an average. Let \(D_F,D_W\) be a random partition of the observations with sizes \(n_F,n_W\), and write \((\mathcal X_F,\mathbf Y_F)\) and \((\mathcal X_W,\mathbf Y_W)\) for the two samples. Candidates are fitted on \(D_F\), while comparisons and weight estimation use \(D_W\).
An admissible support \(S\) is nonempty, satisfies \(|S|\le d\), and has a full-rank fitting design. Let \(\widehat{\boldsymbol\mu}_{S,F}\) denote the vector of predictions on \(\mathcal X_W\) obtained from the OLS fit on \(S\) using \(D_F\); the subscript \(F\) identifies the fitting sample. A variable-selection method tuned within \(D_F\) supplies an admissible initial support \(\widehat S_0\). The initial collection is \(\mathcal C_0=\{\widehat S_0\}\), with weight one assigned to this model.
For a nonempty collection \(\mathcal C\), let \(\mathcal H_n(\mathcal C)=\{\boldsymbol w:w_S\ge0,\ \sum_{S\in\mathcal C}w_S=1\}\) be the unit simplex, and write \(\widehat{\boldsymbol\mu}_F(\boldsymbol w)=\sum_{S\in\mathcal C}w_S\widehat{\boldsymbol\mu}_{S,F}\) for the averaged prediction. The averaging weights minimize the validation criterion
\[
Q(\boldsymbol w;\mathcal C)
=\frac{1}{n_W}
\bigl\|\mathbf Y_W-\widehat{\boldsymbol\mu}_F(\boldsymbol w)\bigr\|^2.
\]
After \(t\) rounds, \(\mathcal C_t\) contains all admitted models, and the current weights \(\widehat{\boldsymbol w}_t\) minimize \(Q\) over \(\mathcal H_n(\mathcal C_t)\).

Each support modification produces a new candidate, while the parent model remains in the collection. The permitted modifications are specified below.

\begin{condition}[Mixed additions, deletions and replacements]
\label{cond:ddp-generation}
Fix integers \(1\le d_v\le d\le p_n\). Every proposed support is admissible and has the form
\[
(S\setminus D)\cup A,\qquad S\in\mathcal C_t,\quad \widehat w_{t,S}>0,\quad D\subseteq S,\quad A\subseteq S^c,\quad |D|,|A|\le d_v.
\]
\end{condition}

Pure additions have \(D=\varnothing\), pure deletions have \(A=\varnothing\), and replacements allow both sets to be nonempty, with possibly different cardinalities. The bound \(d_v\) applies separately to additions and deletions, and no grouping of variables is imposed. The choice of proposals may use the fitting data, current average, search history and independent auxiliary randomization. Residual information can guide which variables to add, delete or replace, but each proposed model is fitted to the original response \(\mathbf Y_F\).

For a proposed support \(S\), define its empirical gain by
\begin{equation*}
\widehat\gamma_t(S)
=Q(\widehat{\boldsymbol w}_t;\mathcal C_t)
-\min_{\boldsymbol w\in\mathcal H_n(\mathcal C_t\cup\{S\})}
Q(\boldsymbol w;\mathcal C_t\cup\{S\}).
\end{equation*}
The gain measures the contribution of \(S\) to the jointly reoptimized average. It is nonnegative because the current average remains feasible, and it equals zero for any support already in \(\mathcal C_t\). The DMA procedure admits a proposal with the largest positive gain and uses the weights from that reoptimization.
Algorithm~\ref{alg:data-driven-path} summarizes the DMA procedure.

\begin{algorithm}[t]
\caption{Dynamic model averaging (DMA)}
\label{alg:data-driven-path}
\begin{algorithmic}[1]
\Require \(D_F,D_W\), an admissible initial support \(\widehat S_0\), dimension bound \(d\), operation size \(d_v\),  round budget \(T\), and a rule for selecting mixed support proposals
\State Fit the initial model on \(D_F\), compute \(\widehat{\boldsymbol\mu}_{\widehat S_0,F}\), and set \(\mathcal C_0=\{\widehat S_0\}\), \(\widehat w_{0,\widehat S_0}=1\)
\For{\(t=0,\ldots,T-1\)}
\State Construct \(\mathcal G_t\) by additions, deletions and replacements satisfying Condition~\ref{cond:ddp-generation}; fit previously unfitted supports on \(D_F\)
\State Compute \(\widehat\gamma_t(S)\) for \(S\in\mathcal G_t\), fully reoptimizing weights for each extension
\If{\(\mathcal G_t\ne\varnothing\) and its largest gain is positive}
\State Choose a maximizing \(S_t^+\) and set \(\mathcal C_{t+1}=\mathcal C_t\cup\{S_t^+\}\)
\State Assign \(\widehat{\boldsymbol w}_{t+1}\) the minimizing weights for this extension
\Else
\State Set \(\mathcal C_{t+1}=\mathcal C_t\) and \(\widehat{\boldsymbol w}_{t+1}=\widehat{\boldsymbol w}_t\)
\EndIf
\EndFor
\Ensure \(\mathcal C_T\) and \(\widehat{\boldsymbol\mu}_T=\widehat{\boldsymbol\mu}_F(\widehat{\boldsymbol w}_T)\)
\end{algorithmic}
\end{algorithm}

For each proposed extension, weight reoptimization is a convex quadratic problem over the unit simplex. The analysis uses exact minimization of the unpenalized validation criterion, with fixed measurable rules for ties and nonunique minimizing weights. At most one model is admitted per round, so \(|\mathcal C_T|\le T+1\). A round without a positive gain leaves the current average unchanged, while subsequent rounds may propose different supports.

\subsection{Prediction risk and unrestricted oracle guarantees}
\label{sec:ddp-risk}

For DMA, candidate construction and weight estimation both use the validation sample. We first quantify the resulting cost of adaptation, then establish conditions under which the generated collection preserves the gains available from unrestricted averaging.
For a deterministic round budget \(T\), write \(R_t=E\|\widehat{\boldsymbol\mu}_t-\boldsymbol\mu_W\|^2/n_W\), \(t=0,\ldots,T\), where \(\widehat{\boldsymbol\mu}_0=\widehat{\boldsymbol\mu}_{\widehat S_0,F}\). Expectations integrate fitting noise, validation noise and auxiliary randomization, conditional on the designs and sample partition. Define the cumulative empirical improvement by \(\widehat{\gamma}_T=Q(\widehat{\boldsymbol w}_0;\mathcal C_0)-Q(\widehat{\boldsymbol w}_T;\mathcal C_T)\ge0\).

\begin{assumption}[Gaussian errors]
\label{ass:ddp-tails}
Conditional on \(\mathcal X\), the regression errors have distribution \(N(\boldsymbol0,\sigma^2I_n)\), with \(0<\sigma^2<\infty\).
\end{assumption}

\begin{theorem}[Empirical improvement and adaptive risk]
\label{thm:ddp-prediction}
Under Assumption~\ref{ass:ddp-tails} and Condition~\ref{cond:ddp-generation}, the DMA estimator satisfies
\begin{equation}
\label{eq:ddp-dynamic-cumulative}
R_T\le R_0-E\widehat{\gamma}_T
+2\sqrt{C\frac{\sigma^2}{n_W}
\min\{p_n,d_vT\log(ep_n)\}\,E\widehat{\gamma}_T},
\end{equation}
where \(C\) is a universal constant. Moreover,
\begin{equation}
\label{eq:ddp-random-hull}
R_T\le E\left[
\min_{\boldsymbol w\in\mathcal H_n(\mathcal C_T)}
\frac{\|\widehat{\boldsymbol\mu}_F(\boldsymbol w)-\boldsymbol\mu_W\|^2}{n_W}
\right]
+C\frac{\sigma^2}{n_W}\min\{p_n,d_vT\log(ep_n)\}.
\end{equation}
\end{theorem}

The first bound relates empirical improvement to risk reduction. If \[\frac{\sigma^2}{n_W}\allowbreak\min\{p_n,d_vT\log(ep_n)\}=o(E\widehat{\gamma}_T),\] then \(R_T\le R_0-\{1-o(1)\}E\widehat{\gamma}_T\). The complexity term accounts for adaptive candidate construction and weight estimation: \(d_vT\log(ep_n)\) reflects the possible support adjustments, while \(p_n\) bounds the dimension of the space containing all fitted predictions. The second bound compares the output with the best combination in the generated collection. We next establish when that collection is sufficiently rich to attain an unrestricted oracle bound.

Suppose that \(\boldsymbol\mu_F=\mathcal X_F\boldsymbol\beta\) and \(\boldsymbol\mu_W=\mathcal X_W\boldsymbol\beta\). Let \(S\) range over all nonempty subsets of \(\{1,\ldots,p_n\}\), and extend the OLS construction in Section~\ref{sec:ddp-construction} to this family. For fixed simplex weights, define \(R_{p_n}(\boldsymbol w)=E\|\widehat{\boldsymbol\mu}_F(\boldsymbol w)-\boldsymbol\mu_W\|^2/n_W\), and let \(\boldsymbol w^*\) minimize this risk. The averaging weights are held fixed in this expectation, and \(\boldsymbol e_S\) assigns unit weight to model \(S\). Thus \(\min_S R_{p_n}(\boldsymbol e_S)\) compares the expected risks of all nonempty subset models, each fitted on \(D_F\) and evaluated on \(\mathcal X_W\).

\begin{assumption}[Orthogonal designs and signal distribution]
\label{ass:ddp-mixed-design}
Suppose that \(\mathcal X_F^T\mathcal X_F=n_FI_{p_n}\), \(\mathcal X_W^T\mathcal X_W=n_WI_{p_n}\), and \(p_n\to\infty\). For fixed constants \(0<c_\beta\le C_\beta<\infty\) and \(0<\pi_0\le1\), assume that
\[
\#\left\{1\le j\le p_n:
c_\beta\le\frac{n_F\beta_j^2}{\sigma^2}\le C_\beta\right\}
\ge\pi_0p_n.
\]
\end{assumption}

The condition places a nonvanishing fraction of the coefficients on the fitting-noise scale. The remaining coefficients may be weaker or stronger, with no uniform upper bound on their standardized magnitudes. The coordinates on the fitting-noise scale provide persistent opportunities for partial inclusion, allowing their individual gains to accumulate into a proportional risk reduction.

Under Assumption~\ref{ass:ddp-mixed-design}, the unrestricted averaging oracle has coordinate inclusion weights \(\frac{\beta_j^2}{\beta_j^2+\sigma^2/n_F}\). For all sufficiently large \(n\),
\begin{equation}
\label{eq:ddp-unrestricted-oracle}
R_{p_n}(\boldsymbol w^*)
=\sum_{j=1}^{p_n}\frac{(\sigma^2/n_F)\beta_j^2}{\beta_j^2+\sigma^2/n_F},
\qquad
\frac{R_{p_n}(\boldsymbol w^*)}{\displaystyle\min_S R_{p_n}(\boldsymbol e_S)}
\le1-\eta_0,
\end{equation}
where \(\eta_0=\pi_0\min\{c_\beta^2/(1+c_\beta),\,1/(1+C_\beta)\}>0\). Both oracle risks have order \(p_n\sigma^2/n_F\). The following theorem transfers this proportional improvement to DMA.

\begin{theorem}[Unrestricted oracle bound and improvement over selection]
\label{thm:ddp-all-subset}
Under Assumptions~\ref{ass:ddp-tails} and~\ref{ass:ddp-mixed-design}, there exist constants \(c_d\in(0,1)\) and \(c_W>1\), depending only on \(c_\beta,C_\beta,\pi_0\), such that the following holds. Suppose that \(d\ge c_dp_n\) and \(n_W\ge c_Wn_F\). Apply Algorithm~\ref{alg:data-driven-path} from any admissible fitting-sample initializer \(\widehat S_0\), comparing every admissible mixed operation from every positive-weight support. If the deterministic round budget satisfies
\[
\frac{d^2}{Td_v^2}
\left(1+\frac{n_F\|\boldsymbol\beta\|^2}{p_n\sigma^2}\right)
\longrightarrow0,
\]
then
\begin{equation}
\label{eq:ddp-all-subset-risk}
R_T\le\{1+o(1)\}R_{p_n}(\boldsymbol w^*),
\qquad
\limsup_{n\to\infty}
\frac{R_T}{\displaystyle\min_S R_{p_n}(\boldsymbol e_S)}\le1-\eta_0.
\end{equation}
\end{theorem}

Theorem~\ref{thm:ddp-all-subset} establishes an unrestricted oracle bound for adaptive candidate construction and weight estimation. The dimension threshold ensures sufficient capacity to represent the unrestricted oracle and the comparison fits used in the risk analysis. The sample-allocation threshold makes the validation cost smaller than the available gain. Both thresholds are fixed, permitting \(d<p_n\) and \(n_W\asymp n_F\). The round condition controls the empirical optimization error relative to \(p_n\sigma^2/n_F\); when the average standardized signal energy is bounded, it reduces to \(Td_v^2/d^2\to\infty\).

The proof uses the coordinates on the fitting-noise scale to obtain an expected improvement over the best fixed-weight average. The model-dimension constraint can be enforced by adjusting inclusion weights on these coordinates alone, allowing the remaining signals to be arbitrarily strong. Validation and optimization errors are then controlled at the same risk scale. The complete mixed comparisons are required for this discovery guarantee, while the preceding adaptive-risk bounds also cover smaller proposal pools. Proofs are provided in Supplement~\ref*{app:ddp-proofs}.

\section{Simulation results}
\label{sec:simulation}

We examine whether DMA's adaptive candidate construction yields predictive gains under different degrees of signal concentration. Motivated by the slow- and fast-decay coefficient comparisons in \citet{peng2022improvability} and \citet{xu2025model}, we generate \(Y_i=\mathbf{X}_i^T\bm\beta+\epsilon_i\), where \(\mathbf{X}_i\sim N_p(0,\Sigma)\), \(\Sigma_{jk}=\rho^{|j-k|}\), and \(\epsilon_i\sim N(0,\sigma^2)\) independently. We consider polynomial and exponential decay of the regression coefficients, specified by \(\beta_j=j^{-\alpha}\) and  \(\beta_j=e^{-\alpha j}\), with \(\alpha\in\{0.5,1\}\), \((n,p)\in\{(100,64),(200,102)\}\), and \(\rho\in\{0,0.3,0.5\}\). The noise variance is chosen so that the population \(R^2\) ranges from \(0.1\) to \(0.9\).

\begin{figure}[tbp]
\centering
\includegraphics[width=14.2cm]{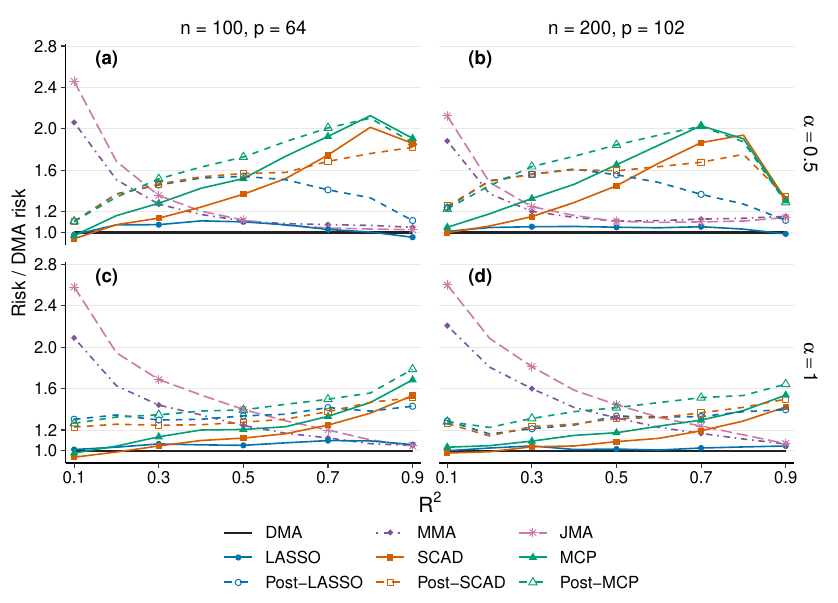}
\caption{Estimated prediction risks relative to LASSO-initialized DMA under polynomial decay, with \(\rho=0.3\). Rows correspond to \(\alpha=0.5\) and \(1\); columns correspond to \((n,p)=(100,64)\) and \((200,102)\). Each risk is averaged over 100 replications. Ratios above one favor DMA.}
\label{fig:simulation-polynomial}
\end{figure}

\begin{figure}[tbp]
\centering
\includegraphics[width=14.2cm]{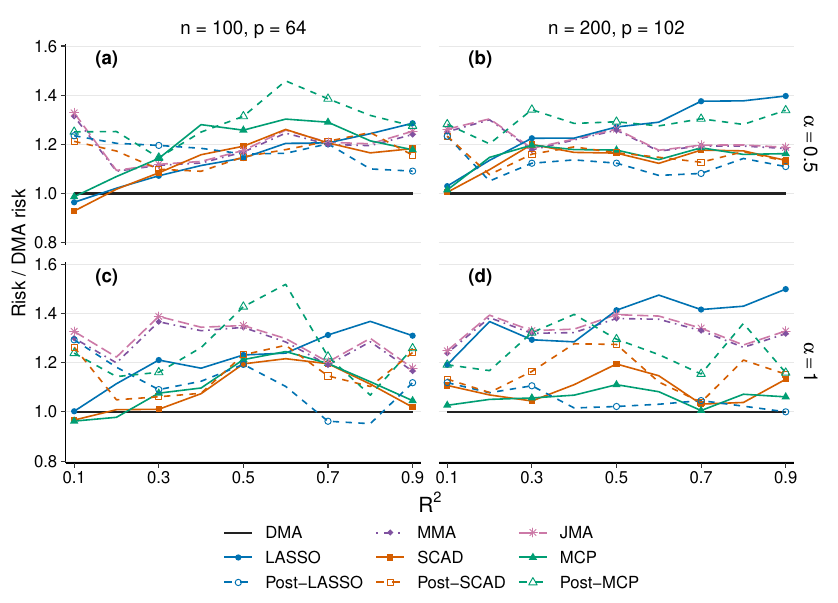}
\caption{Estimated prediction risks relative to Post-LASSO-initialized DMA under exponential decay, with \(\rho=0.3\). Rows correspond to \(\alpha=0.5\) and \(1\); columns correspond to \((n,p)=(100,64)\) and \((200,102)\). Each risk is averaged over 100 replications. Ratios above one favor DMA.}
\label{fig:simulation-exponential}
\end{figure}

We compare DMA with LASSO \citep{tibshirani1996}, SCAD \citep{fan2001variable}, MCP \citep{zhang2010minimax}, and their post-selection least-squares counterparts, denoted by Post-LASSO, Post-SCAD and Post-MCP. Each post-selection procedure refits the models along the corresponding penalized solution path by ordinary least squares and selects its tuning parameter separately by minimizing the five-fold cross-validated mean squared error of the refits; see \citet{belloni2013postselection} for the Post-LASSO formulation. We also include nested MMA \citep{hansen2007least} and JMA \citep{hansen2012jackknife}, with candidate models constructed according to the order in which variables enter the LASSO solution path under polynomial decay and the MCP solution path under exponential decay. For DMA, the choice of initialization reflects the different roles of shrinkage under the two decay regimes. Under polynomial decay, we use LASSO initialization because retaining shrinkage helps stabilize estimation of a signal distributed across many coefficients. Under exponential decay, we use Post-LASSO initialization because, when a small selected model captures most of the signal, least-squares refitting can reduce shrinkage bias in the dominant coefficients without an excessive increase in variance. Each choice is held fixed across all settings within the corresponding decay regime. We evaluate predictive performance using prediction risk averaged over 100 replications. Figures~\ref{fig:simulation-polynomial} and~\ref{fig:simulation-exponential} report each method's average risk relative to that of DMA for \(\rho=0.3\); ratios above one favor DMA. Details of nested candidate construction, tuning and sample splitting, together with comparisons with alternative initializations and results for the remaining correlations, are provided in Supplement~\ref*{app:simulation}.

Under polynomial decay, DMA improves over LASSO in most displayed settings and generally achieves larger gains over post-selection least squares and nested MMA and JMA (Figure~\ref{fig:simulation-polynomial}). These comparisons reflect the difficulty of estimating a dispersed signal: omitting many weak coefficients accumulates approximation error, whereas retaining them without shrinkage increases estimation variability. Averaging allows predictors represented in different candidates to contribute partially, balancing these costs. LASSO provides a particularly competitive benchmark because it also combines variable retention with shrinkage. The smaller gains over LASSO than over Post-LASSO therefore underscore the importance of controlling estimation variance alongside recovering signal.
The comparisons with MMA and JMA further highlight the role of the candidate model set. Both methods optimize weights within a prescribed collection, so their predictive performance depends on which models that collection contains. Along a nested sequence, including a later-entering predictor requires retaining all preceding predictors, which restricts the available combinations of weak signals and estimation costs. DMA constructs candidates according to their contribution to the current average, using additions, deletions and replacements to explore complementary supports. Its lower risks relative to the specified MMA and JMA implementations illustrate the predictive value of dynamic candidate construction together with joint weight reoptimization.

Under exponential decay, DMA with Post-LASSO initialization improves on Post-LASSO in most displayed settings and has lower estimated risk than both nested MMA and JMA throughout (Figure~\ref{fig:simulation-exponential}). Signal concentration changes the relevant tradeoff: once a small model captures the dominant predictors, accurate estimation of their coefficients becomes particularly valuable. LASSO uses the same penalty to control variable inclusion and shrink retained coefficients, so effective screening can still introduce appreciable bias in the dominant effects. Least-squares refitting removes the explicit penalization of these coefficients, while SCAD and MCP reduce penalization as coefficient magnitude increases. The stronger performance of these methods, particularly when \(\alpha=1\), is consistent with the increased value of reducing shrinkage bias when the signal is concentrated.
A good sparse approximation nevertheless leaves uncertainty about which weaker predictors should be included. DMA can address this uncertainty through support adjustments and averaging, yielding further gains beyond the Post-LASSO initializer. These gains vary with the remaining approximation error and the statistical costs of constructing candidates and estimating their weights. For MMA and JMA, signal concentration alone does not ensure that the nested collection contains suitable approximations: the solution-path ordering and retained model sizes still determine the available supports. DMA can revise the supports suggested by its initializer and assess new candidates jointly with those already admitted. Its advantage over the specified nested procedures indicates that candidate construction remains consequential even under rapid decay. This finite-sample comparison complements the oracle analysis: a limited asymptotic gain over the selection oracle can coexist with useful improvements over procedures whose candidate models and weights must be estimated.

When the degree of signal concentration is unknown, both LASSO and Post-LASSO can be included in the initial candidate collection, with their combination estimated on the weighting sample before further candidate expansion. Results for this joint initialization, together with comparisons with other initializations, are reported in Supplement~\ref*{app:simulation}.

\FloatBarrier
\section{Real-data analysis}
\label{sec:real-data}

We examine predictive performance on 32 regression data sets with \(n<4500\) and \(p<1000\), drawn from published studies of model selection and regularized regression and publicly available data collections \citep{dataGuide,dataHitters,dataSolubility}. The collection covers molecular properties, socioeconomic outcomes, spectroscopy, gene expression and functional measurements. Table~\ref{tab:real-data-sets} summarizes the data sets, their sources, sample sizes and predictor dimensions. It comprises 20 low-dimensional data sets with \(p<n\) and 12 high-dimensional data sets with \(p\ge n\). We report results separately for these two groups to examine how the magnitude and prevalence of predictive gains vary across dimensional regimes.

\begin{table}[tbp]
\centering
\begin{threeparttable}
\caption{Data sets used in the empirical comparison.}
\label{tab:real-data-sets}
\small
\setlength{\tabcolsep}{3pt}
\renewcommand{\arraystretch}{1.12}
\begin{tabular*}{\textwidth}{@{\extracolsep{\fill}}p{0.34\textwidth}r p{0.34\textwidth}r@{}}
\toprule
Data set and reference & \((n,p)\) & Data set and reference & \((n,p)\) \\
\midrule
\multicolumn{4}{@{}l}{\textit{Low-dimensional: \(p<n\)}} \\
\addlinespace[2pt]
Prostate \citep{hastie2009elements} & \((97,8)\) & Wage1 \citep{hansen2012jackknife} & \((526,29)\) \\
Fifth National Bank \citep{fan2004diverging} & \((208,25)\) & FreeSolv \citep{mobley2014freesolv} & \((642,217)\) \\
CPU \citep{dataGuide} & \((209,70)\) & College \citep{dataGuide} & \((694,70)\) \\
Tecator \citep{dataTecator} & \((215,100)\) & B3DB \citep{meng2021b3db} & \((1058,217)\) \\
Bodyfat \citep{johnson1996bodyfat} & \((252,13)\) & VDss \citep{lombardo2016vdss} & \((1130,217)\) \\
Hitters \citep{dataHitters} & \((263,19)\) & Solubility \citep{dataSolubility} & \((1267,228)\) \\
Ozone \citep{dataGuide} & \((330,32)\) & BACE \citep{subramanian2016bace} & \((1513,589)\) \\
Auto MPG \citep{dataGuide} & \((392,8)\) & Budget \citep{dataGuide} & \((1729,60)\) \\
Diabetes \citep{efron2004leastangle} & \((442,64)\) & Amenity \citep{dataGuide} & \((3044,70)\) \\
Boston Housing \citep{harrison1978hedonic} & \((506,91)\) & Abalone \citep{dataGuide} & \((4177,9)\) \\
\midrule
\multicolumn{4}{@{}l}{\textit{High-dimensional: \(p\ge n\)}} \\
\addlinespace[2pt]
Yarn \citep{polson2012local} & \((28,268)\) & Corn \citep{dataCorn} & \((80,700)\) \\
Canadian Weather \citep{james2009functional} & \((35,365)\) & Soil \citep{lin2017ohpl} & \((108,700)\) \\
Nutrimouse \citep{polson2012local} & \((40,120)\) & Eye200 \citep{li2015flare} & \((120,200)\) \\
Gasoline \citep{polson2012local} & \((60,401)\) & Fermentation \citep{liebmann2009fermentation} & \((166,235)\) \\
Multidrug \citep{polson2012local} & \((60,853)\) & Orange juice \citep{thind2020functional} & \((218,700)\) \\
Biscuit \citep{dataBiscuit} & \((72,700)\) & Sugar \citep{bro1999sugar} & \((268,571)\) \\
\bottomrule
\end{tabular*}
\end{threeparttable}
\end{table}

 We compare DMA and its shrinkage version, s-DMA, with the six MS methods considered in Section~\ref{sec:simulation}. To limit estimation instability in the high-dimensional experiments, least-squares candidates and post-selection refits use full-rank designs with at most \(\lfloor n_{\mathrm{fit}}/3\rfloor\) predictors, where \(n_{\mathrm{fit}}\) denotes the number of observations used in the corresponding least-squares fit.
Both DMA and s-DMA use LASSO initialization for every data set. The shrinkage version regularizes the averaged prediction toward this initial estimate. Specifically, s-DMA minimizes
\[
Q_{\lambda}(\boldsymbol w;\mathcal C)
=
\frac{1}{n_W}
\left\|
\boldsymbol Y_W-\widehat{\boldsymbol\mu}_{F}(\boldsymbol w)
\right\|^2
+
\frac{\lambda}{n_W}
\left\|
\widehat{\boldsymbol\mu}_{F}(\boldsymbol w)
-\widehat{\boldsymbol\mu}_{0,F}
\right\|^2,
\qquad
\boldsymbol w\in\mathcal H_n(\mathcal C),
\]
where \(\widehat{\boldsymbol\mu}_{0,F}\) denotes the LASSO prediction on the weighting sample and remains fixed throughout the search. The case \(\lambda=0\) gives DMA. For \(\lambda>0\), the additional term discourages large departures from this initial prediction.

For each data set, we conduct 100 repetitions using 80\% of the observations for training and 20\% for testing. All methods use the same training sample. DMA partitions this sample into three disjoint subsets for candidate fitting, weight estimation and tuning, rotates the partition ten times, and averages the resulting predictions. The same partitions are used for s-DMA. Within each rotation, s-DMA jointly selects the path length and \(\lambda\in\{0,1/9,\ldots,1\}\) by minimizing prediction MSE on the tuning subset. The MS methods are tuned by five-fold cross-validation within the training sample. 

\FloatBarrier
\subsection{Predictive performance}
\label{sec:real-data-performance}

For a test sample of size \(n_{\mathrm{test}}\), prediction error is measured by \(\operatorname{MSE}=n_{\mathrm{test}}^{-1}\sum_{i=1}^{n_{\mathrm{test}}}\{Y_i-\widehat f(X_i)\}^{2}\). We first average the test MSE over the 100 repetitions for each method and data set. The percentage improvement over an MS method is then calculated as \(100(1-\operatorname{MSE}_{\mathrm{DMA}}/\operatorname{MSE}_{\mathrm{MS}})\), with DMA replaced by s-DMA for the shrinkage version, where both MSE values denote these repetition averages. Table~\ref{tab:real-data-improvement} reports the median and the 25th and 75th percentiles of the resulting percentages within each dimensional group, together with the number of data sets favoring the averaging procedure.

\begin{table}[htbp]
\centering
\begin{threeparttable}
\caption{Median percentage reductions in test MSE relative to the MS comparison methods. Parentheses give the 25th--75th percentiles. Brackets report the number of data sets favoring the averaging procedure out of the group total.}
\label{tab:real-data-improvement}
\small
\setlength{\tabcolsep}{3pt}
\renewcommand{\arraystretch}{1.15}
\begin{tabular*}{\textwidth}{@{\extracolsep{\fill}}llrrrrrr@{}}
\toprule
Dimensional regime & Method & LASSO & SCAD & MCP & \shortstack{Post-\\LASSO} & \shortstack{Post-\\SCAD} & \shortstack{Post-\\MCP} \\
\midrule
Low-dimensional & DMA & \shortstack[r]{0.47\\{\scriptsize (0.07--2.49)}\\{\scriptsize [16/20]}} & \shortstack[r]{5.50\\{\scriptsize (1.20--11.40)}\\{\scriptsize [20/20]}} & \shortstack[r]{6.03\\{\scriptsize (1.14--12.39)}\\{\scriptsize [19/20]}} & \shortstack[r]{3.81\\{\scriptsize (1.92--16.90)}\\{\scriptsize [20/20]}} & \shortstack[r]{4.80\\{\scriptsize (1.46--11.22)}\\{\scriptsize [20/20]}} & \shortstack[r]{5.61\\{\scriptsize (1.45--11.22)}\\{\scriptsize [20/20]}} \\
(20 data sets) & s-DMA & \shortstack[r]{0.47\\{\scriptsize (0.15--2.62)}\\{\scriptsize [17/20]}} & \shortstack[r]{5.72\\{\scriptsize (1.25--11.58)}\\{\scriptsize [20/20]}} & \shortstack[r]{6.31\\{\scriptsize (1.18--12.56)}\\{\scriptsize [19/20]}} & \shortstack[r]{4.10\\{\scriptsize (2.02--17.04)}\\{\scriptsize [20/20]}} & \shortstack[r]{4.93\\{\scriptsize (1.44--11.40)}\\{\scriptsize [20/20]}} & \shortstack[r]{5.88\\{\scriptsize (1.51--11.40)}\\{\scriptsize [20/20]}} \\
\addlinespace
High-dimensional & DMA & \shortstack[r]{7.50\\{\scriptsize (1.92--19.28)}\\{\scriptsize [11/12]}} & \shortstack[r]{15.50\\{\scriptsize (9.52--35.00)}\\{\scriptsize [11/12]}} & \shortstack[r]{19.93\\{\scriptsize (10.60--41.50)}\\{\scriptsize [11/12]}} & \shortstack[r]{13.73\\{\scriptsize (6.79--55.92)}\\{\scriptsize [12/12]}} & \shortstack[r]{16.17\\{\scriptsize (10.92--34.27)}\\{\scriptsize [11/12]}} & \shortstack[r]{20.65\\{\scriptsize (9.85--41.25)}\\{\scriptsize [11/12]}} \\
(12 data sets) & s-DMA & \shortstack[r]{7.17\\{\scriptsize (4.40--19.14)}\\{\scriptsize [11/12]}} & \shortstack[r]{15.14\\{\scriptsize (10.08--35.27)}\\{\scriptsize [11/12]}} & \shortstack[r]{19.21\\{\scriptsize (10.60--43.52)}\\{\scriptsize [11/12]}} & \shortstack[r]{12.94\\{\scriptsize (6.51--57.88)}\\{\scriptsize [12/12]}} & \shortstack[r]{15.52\\{\scriptsize (10.39--34.55)}\\{\scriptsize [11/12]}} & \shortstack[r]{20.56\\{\scriptsize (10.14--43.25)}\\{\scriptsize [11/12]}} \\
\bottomrule
\end{tabular*}
\begin{tablenotes}[flushleft]
\footnotesize
\item[] \textit{Notes.} Both averaging procedures use LASSO initialization. Positive entries indicate lower test MSE for the corresponding dynamic averaging procedure.
\end{tablenotes}
\end{threeparttable}
\end{table}

\begin{figure}[htbp]
\centering
\includegraphics[width=\textwidth]{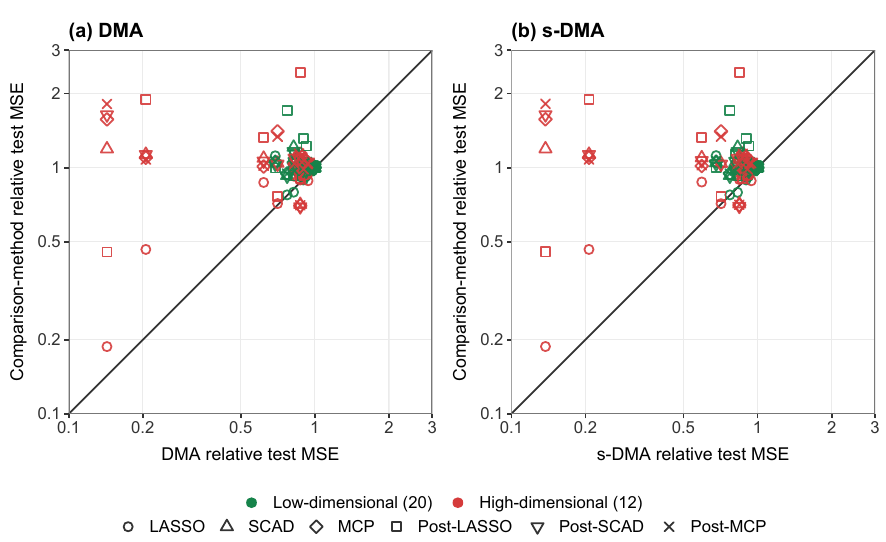}
\caption{Pairwise comparisons of normalized mean test MSE for DMA (left) and s-DMA (right), both initialized by LASSO. Green and red indicate low- and high-dimensional data sets, respectively; plotting symbols identify the six MS comparison methods. Points above the diagonal favor the averaging procedure in the corresponding panel.}
\label{fig:real-data-comparison}
\end{figure}

Both averaging procedures achieve positive median improvements against all six MS methods in both dimensional groups (Table~\ref{tab:real-data-improvement}). The gains also extend across data sets: DMA and s-DMA achieve lower mean test MSE than all six MS methods simultaneously on 26 and 27 of the 32 data sets, respectively. Figure~\ref{fig:real-data-comparison} displays the comparisons for individual data sets using DMA in the left panel and s-DMA in the right panel. To remove differences in response scale, within each panel we divide the mean test MSEs of the averaging procedure and the six MS methods by their common arithmetic mean within each data set. The horizontal coordinate gives the normalized MSE of the averaging procedure and the vertical coordinate that of an MS method. Points above the diagonal favor the averaging procedure, and both axes use logarithmic scales. 

In the low-dimensional group, the median reductions relative to LASSO are 0.47\% for both procedures, whereas those relative to the other five MS methods range from 3.81\% to 6.31\%. The smaller gains over LASSO are also reflected in Figure~\ref{fig:real-data-comparison}, where these comparisons tend to lie close to the diagonal. These modest gains are consistent with more limited scope for improvement when estimation variability and uncertainty about the selected support are less consequential. Moreover, test MSE includes irreducible response variation, so a reduction in estimation error may translate into only a small percentage reduction in total prediction error. LASSO provides a particularly competitive comparison because it combines variable selection with coefficient shrinkage. Retaining weak or correlated predictors with reduced coefficients can preserve predictive information while moderating estimation variability, providing some of the same benefits sought through averaging. Post-selection least squares removes this shrinkage and may increase variance. Similarly, the reduced shrinkage of large coefficients under SCAD and MCP can be valuable when shrinkage bias is consequential, but need not improve prediction when variance control is more important. These considerations offer a possible explanation for the smaller gains over LASSO than over the other MS methods.

In the high-dimensional group, the median reductions relative to LASSO increase to 7.50\% for DMA and 7.17\% for s-DMA, with both procedures outperforming LASSO on 11 of the 12 data sets. Against the other five MS methods, the median reductions range from 12.94\% to 20.65\%; both averaging procedures outperform Post-LASSO on all 12 data sets and each of the remaining four methods on 11. These larger gains are consistent with greater scope for balancing information retention against estimation variability. When many predictors are available relative to the sample size, several competing supports may fit the available data similarly while retaining different predictive information. Selecting a single support can omit useful directions represented in other candidates, whereas fitting a larger model to include those directions may incur substantial variance. DMA can combine complementary candidates and allow their predictors to contribute through the averaging weights. Each least-squares candidate remains restricted in size, while the combined predictor can retain information from a broader collection of predictors without fitting an unrestricted least-squares model on their union.
The dynamic construction of the candidate collection can further support this balance. Through support adjustments and joint weight reoptimization, DMA can retain a candidate for its contribution to the combined prediction, even when that candidate would not be preferred on its own. The potential gain depends on whether the candidates provide complementary predictive information and whether their weights can be estimated reliably; different supports alone do not ensure an improvement. The high-dimensional results are consistent with greater opportunities for such complementarity.

\section{Conclusion and discussion} \label{sec:: concolusion}

This paper contributes to both the theoretical understanding of non-nested least-squares model averaging and the methodology for constructing candidate model collections from data. Our analysis and methodological development are united by a iterative principle: candidates are evaluated by their predictive gain after jointly reoptimizing the averaging weights over the enlarged collection, so that the model list and the averaging weights evolve together.

From a theoretical perspective, our analysis of oracle performance clarifies when non-nested averaging moves beyond the one-half limitation of nested averaging to achieve same-order or order-improvement over selection, driven jointly by RMAI and projection geometry. We also prove that a small collection of bounded dimension non-nested models can preserve the gains of all-subset averaging by combining complementary subsets of predictors under both orthogonal and correlated designs. 
 From a methodological perspective, DMA turns these insights into an adaptive procedure that expands the candidate collection in response to the evolving average while controlling the statistical cost of reusing validation data. In both simulations and real-data applications, this dynamic construction proves most valuable in high-dimensional settings, where multiple competing supports carry complementary predictive information.
 
 Several open questions need further investigation. First, our all-subset oracle guarantee for DMA relies on orthogonal designs and benchmarks performance based on the fitting subsample rather than the full sample; extending this guarantee to general correlated designs and to full-sample benchmarks remains an important theoretical challenge. Second, our correlated-design oracle results rely on inverse-Gram diagonal dominance, whereas many high-dimensional econometric applications exhibit strong collinearity; understanding how non-nested averaging retains its gains under strongly correlated or factor-structured designs remains an open theoretical question.
  Finally, although our theoretical analysis relies on the projection geometry of linear least squares, the iterative principle of adaptively modifying candidates and jointly reoptimizing averaging weights extends naturally to nonparametric and nonlinear learners, such as penalized splines, additive models, and tree ensembles.
\bibliographystyle{plainnat}
\bibliography{bibl1}
\newpage
\end{document}

%% file: MA.bbl
\begin{thebibliography}{62}
\providecommand{\natexlab}[1]{#1}
\providecommand{\url}[1]{\texttt{#1}}
\expandafter\ifx\csname urlstyle\endcsname\relax
  \providecommand{\doi}[1]{doi: #1}\else
  \providecommand{\doi}{doi: \begingroup \urlstyle{rm}\Url}\fi

\bibitem[Ando and Li(2014)]{ando2014model}
Tomohiro Ando and Ker-Chau Li.
\newblock A model-averaging approach for high-dimensional regression.
\newblock \emph{Journal of the American Statistical Association}, 109\penalty0
  (505):\penalty0 254--265, 2014.

\bibitem[{APM}(n.d.)]{dataSolubility}
{APM}.
\newblock Solubility data in the {AppliedPredictiveModeling} package.
\newblock \url{https://github.com/cran/AppliedPredictiveModeling}, n.d.

\bibitem[Baraud et~al.(2009)Baraud, Giraud, and Huet]{baraud2009gaussian}
Yannick Baraud, Christophe Giraud, and Sylvie Huet.
\newblock Gaussian model selection with an unknown variance.
\newblock \emph{The Annals of Statistics}, 37\penalty0 (2):\penalty0 630--672,
  2009.
\newblock \doi{10.1214/07-AOS573}.

\bibitem[Barron et~al.(1999)Barron, Birg{\'e}, and Massart]{barron1999risk}
Andrew Barron, Lucien Birg{\'e}, and Pascal Massart.
\newblock Risk bounds for model selection via penalization.
\newblock \emph{Probability Theory and Related Fields}, 113\penalty0
  (3):\penalty0 301--413, 1999.
\newblock \doi{10.1007/s004400050210}.

\bibitem[Bellec(2016)]{bellec2016aggregation}
Pierre~C. Bellec.
\newblock Aggregation of supports along the {Lasso} path.
\newblock In \emph{Proceedings of the 29th Annual Conference on Learning
  Theory}, volume~49 of \emph{Proceedings of Machine Learning Research}, pages
  488--529. PMLR, 2016.
\newblock URL \url{https://proceedings.mlr.press/v49/bellec16.html}.

\bibitem[Bellec(2018)]{bellec2018optimal}
Pierre~C Bellec.
\newblock Optimal bounds for aggregation of affine estimators.
\newblock \emph{The Annals of Statistics}, 46\penalty0 (1):\penalty0 30--59,
  2018.

\bibitem[Belloni and Chernozhukov(2013)]{belloni2013postselection}
Alexandre Belloni and Victor Chernozhukov.
\newblock Least squares after model selection in high-dimensional sparse
  models.
\newblock \emph{Bernoulli}, 19\penalty0 (2):\penalty0 521--547, 2013.
\newblock \doi{10.3150/11-BEJ410}.

\bibitem[Birg{\'e} and Massart(2001)]{birge2001gaussian}
Lucien Birg{\'e} and Pascal Massart.
\newblock Gaussian model selection.
\newblock \emph{Journal of the European Mathematical Society}, 3\penalty0
  (3):\penalty0 203--268, 2001.
\newblock \doi{10.1007/s100970100031}.

\bibitem[Bro(1999)]{bro1999sugar}
Rasmus Bro.
\newblock Exploratory study of sugar production using fluorescence spectroscopy
  and multi-way analysis.
\newblock \emph{Chemometrics and Intelligent Laboratory Systems}, 1999.
\newblock URL \url{https://ucphchemometrics.com/sugar-process-data/}.

\bibitem[Bu et~al.(2025)Bu, Liang, Zhang, and Zou]{bu2025improving}
Qiushi Bu, Hua Liang, Xinyu Zhang, and Jiahui Zou.
\newblock Improving tensor regression by optimal model averaging.
\newblock \emph{Journal of the American Statistical Association}, 120\penalty0
  (550):\penalty0 1115--1126, 2025.
\newblock \doi{10.1080/01621459.2024.2398164}.

\bibitem[Dalalyan and Tsybakov(2008)]{dalalyan2008aggregation}
Arnak Dalalyan and Alexandre~B Tsybakov.
\newblock Aggregation by exponential weighting, sharp {PAC-Bayesian} bounds and
  sparsity.
\newblock \emph{Machine Learning}, 72\penalty0 (1--2):\penalty0 39--61, 2008.

\bibitem[Dalalyan and Salmon(2012)]{dalalyan2012sharp}
Arnak~S Dalalyan and Joseph Salmon.
\newblock Sharp oracle inequalities for aggregation of affine estimators.
\newblock \emph{The Annals of Statistics}, 40\penalty0 (4):\penalty0
  2327--2355, 2012.

\bibitem[Ding et~al.(2018)Ding, Tarokh, and Yang]{ding2018model}
Jie Ding, Vahid Tarokh, and Yuhong Yang.
\newblock Model selection techniques: An overview.
\newblock \emph{IEEE Signal Processing Magazine}, 35\penalty0 (6):\penalty0
  16--34, 2018.

\bibitem[Duan et~al.(2012)Duan, Soussen, Brie, Idier, and Wang]{duan2012lars}
Junbo Duan, Charles Soussen, David Brie, J{\'e}r{\^o}me Idier, and Yu-Ping
  Wang.
\newblock On {LARS}/homotopy equivalence conditions for over-determined
  {LASSO}.
\newblock \emph{IEEE Signal Processing Letters}, 19\penalty0 (12):\penalty0
  894--897, 2012.
\newblock \doi{10.1109/LSP.2012.2221712}.

\bibitem[Efron et~al.(2004)Efron, Hastie, Johnstone, and
  Tibshirani]{efron2004leastangle}
Bradley Efron, Trevor Hastie, Iain Johnstone, and Robert Tibshirani.
\newblock Least angle regression.
\newblock \emph{The Annals of Statistics}, 32\penalty0 (2):\penalty0 407--499,
  2004.
\newblock \doi{10.1214/009053604000000067}.

\bibitem[{Eigenvector Research}(n.d.)]{dataCorn}
{Eigenvector Research}.
\newblock Corn near-infrared spectroscopy data.
\newblock \url{https://www.eigenvector.com/data/Corn/corn.mat}, n.d.

\bibitem[Fan and Li(2001)]{fan2001variable}
Jianqing Fan and Runze Li.
\newblock Variable selection via nonconcave penalized likelihood and its oracle
  properties.
\newblock \emph{Journal of the American statistical Association}, 96\penalty0
  (456):\penalty0 1348--1360, 2001.

\bibitem[Fan and Peng(2004)]{fan2004diverging}
Jianqing Fan and Heng Peng.
\newblock Nonconcave penalized likelihood with a diverging number of
  parameters.
\newblock \emph{The Annals of Statistics}, 32\penalty0 (3):\penalty0 928--961,
  2004.
\newblock \doi{10.1214/009053604000000256}.

\bibitem[{fda.usc}(n.d.)]{dataTecator}
{fda.usc}.
\newblock Tecator data.
\newblock \url{https://github.com/cran/fda.usc/blob/master/man/tecator.Rd},
  n.d.

\bibitem[{fds}(n.d.)]{dataBiscuit}
{fds}.
\newblock Biscuit dough piece data in {Functional Data Sets}.
\newblock \url{https://github.com/cran/fds}, n.d.

\bibitem[Hansen(2007)]{hansen2007least}
Bruce~E Hansen.
\newblock Least squares model averaging.
\newblock \emph{Econometrica}, 75\penalty0 (4):\penalty0 1175--1189, 2007.

\bibitem[Hansen and Racine(2012)]{hansen2012jackknife}
Bruce~E Hansen and Jeffrey~S Racine.
\newblock Jackknife model averaging.
\newblock \emph{Journal of Econometrics}, 167\penalty0 (1):\penalty0 38--46,
  2012.

\bibitem[Harrison and Rubinfeld(1978)]{harrison1978hedonic}
David Harrison and Daniel~L. Rubinfeld.
\newblock Hedonic housing prices and the demand for clean air.
\newblock \emph{Journal of Environmental Economics and Management}, 5\penalty0
  (1):\penalty0 81--102, 1978.
\newblock \doi{10.1016/0095-0696(78)90006-2}.

\bibitem[Hastie et~al.(2009)Hastie, Tibshirani, and
  Friedman]{hastie2009elements}
Trevor Hastie, Robert Tibshirani, and Jerome Friedman.
\newblock \emph{The Elements of Statistical Learning}.
\newblock Springer, 2 edition, 2009.

\bibitem[{ISLR}(n.d.)]{dataHitters}
{ISLR}.
\newblock Hitters: Major league baseball data.
\newblock \url{https://github.com/cran/ISLR/blob/master/man/Hitters.Rd}, n.d.

\bibitem[James et~al.(2009)James, Wang, and Zhu]{james2009functional}
Gareth~M. James, Jing Wang, and Ji~Zhu.
\newblock Functional linear regression that's interpretable.
\newblock \emph{The Annals of Statistics}, 37\penalty0 (5A):\penalty0
  2083--2108, 2009.
\newblock URL \url{https://arxiv.org/abs/0908.2918}.

\bibitem[Johnson(1996)]{johnson1996bodyfat}
Roger~W. Johnson.
\newblock Fitting percentage of body fat to simple body measurements.
\newblock \emph{Journal of Statistics Education}, 4\penalty0 (1), 1996.

\bibitem[Lan et~al.(2018)Lan, Ma, Zhao, Wang, and Tsai]{lan2018sequential}
Wei Lan, Yingying Ma, Junlong Zhao, Hansheng Wang, and Chih-Ling Tsai.
\newblock Sequential model averaging for high dimensional linear regression
  models.
\newblock \emph{Statistica Sinica}, 28\penalty0 (1):\penalty0 449--469, 2018.
\newblock \doi{10.5705/ss.202016.0122}.

\bibitem[Leung and Barron(2006)]{leung2006information}
Gilbert Leung and Andrew~R Barron.
\newblock Information theory and mixing least-squares regressions.
\newblock \emph{IEEE Transactions on Information Theory}, 52\penalty0
  (8):\penalty0 3396--3410, 2006.

\bibitem[Li(1987)]{li1987asymptotic}
Ker-Chau Li.
\newblock Asymptotic optimality for cp, cl, cross-validation and generalized
  cross-validation: discrete index set.
\newblock \emph{The Annals of Statistics}, pages 958--975, 1987.

\bibitem[Li et~al.(2015)Li, Zhao, Yuan, and Liu]{li2015flare}
Xingguo Li, Tuo Zhao, Xiaoming Yuan, and Han Liu.
\newblock The {flare} package for high dimensional linear regression and
  precision matrix estimation in {R}.
\newblock \emph{Journal of Machine Learning Research}, 16:\penalty0 553--557,
  2015.
\newblock URL \url{https://www.jmlr.org/papers/volume16/li15a/li15a.pdf}.

\bibitem[Liebmann et~al.(2009)Liebmann, Friedl, and
  Varmuza]{liebmann2009fermentation}
B.~Liebmann, A.~Friedl, and K.~Varmuza.
\newblock Determination of glucose and ethanol in bioethanol production by near
  infrared spectroscopy and chemometrics.
\newblock \emph{Analytica Chimica Acta}, 642:\penalty0 171--178, 2009.
\newblock \doi{10.1016/j.aca.2008.10.069}.

\bibitem[Lin et~al.(2017)Lin, Xiao, Wang, Li, and Xu]{lin2017ohpl}
You-Wu Lin, Nan Xiao, Li-Li Wang, Chuan-Quan Li, and Qing-Song Xu.
\newblock Ordered homogeneity pursuit lasso for group variable selection with
  applications to spectroscopic data.
\newblock \emph{Chemometrics and Intelligent Laboratory Systems}, 168:\penalty0
  62--71, 2017.
\newblock \doi{10.1016/j.chemolab.2017.07.004}.

\bibitem[Loh(n.d.)]{dataGuide}
Wei-Yin Loh.
\newblock {GUIDE} regression data archive.
\newblock
  \url{https://pages.stat.wisc.edu/~loh/treeprogs/datasets/Reg-data.zip}, n.d.

\bibitem[Lombardo and Jing(2016)]{lombardo2016vdss}
F.~Lombardo and Y.~Jing.
\newblock In silico prediction of volume of distribution in humans: Extensive
  data set and the exploration of linear and nonlinear methods coupled with
  molecular interaction fields descriptors.
\newblock \emph{Journal of Chemical Information and Modeling}, 2016.
\newblock \doi{10.1021/acs.jcim.6b00044}.

\bibitem[Meinshausen et~al.(2007)Meinshausen, Rocha, and
  Yu]{meinshausen2007discussion}
Nicolai Meinshausen, Guilherme Rocha, and Bin Yu.
\newblock Discussion: A tale of three cousins: {Lasso}, {$L_2$Boosting} and
  {Dantzig}.
\newblock \emph{The Annals of Statistics}, 35\penalty0 (6):\penalty0
  2373--2384, 2007.
\newblock \doi{10.1214/009053607000000460}.

\bibitem[Meng et~al.(2021)Meng, Xi, Huang, and Ayers]{meng2021b3db}
Fanwang Meng, Yang Xi, Jinfeng Huang, and Paul~W. Ayers.
\newblock A curated diverse molecular database of blood--brain barrier
  permeability with chemical descriptors.
\newblock \emph{Scientific Data}, 2021.
\newblock URL \url{https://github.com/theochem/B3DB}.

\bibitem[Mobley and Guthrie(2014)]{mobley2014freesolv}
David~L. Mobley and J.~Peter Guthrie.
\newblock {FreeSolv}: A database of experimental and calculated hydration free
  energies, with input files.
\newblock \emph{Journal of Computer-Aided Molecular Design}, 28:\penalty0
  711--720, 2014.
\newblock \doi{10.1007/s10822-014-9747-x}.

\bibitem[Peng(2025)]{peng2025mallows}
Jingfu Peng.
\newblock Mallows-type model averaging: Non-asymptotic analysis and all-subset
  combination.
\newblock \emph{arXiv preprint arXiv:2505.02637}, 2025.

\bibitem[Peng(2026)]{peng2026optimalrates}
Jingfu Peng.
\newblock Optimal rates for aggregation of affine estimators.
\newblock \emph{arXiv preprint arXiv:2609.07166}, 2026.
\newblock \doi{10.48550/arXiv.2609.07166}.

\bibitem[Peng and Yang(2022)]{peng2022improvability}
Jingfu Peng and Yuhong Yang.
\newblock On improvability of model selection by model averaging.
\newblock \emph{Journal of Econometrics}, 229\penalty0 (2):\penalty0 246--262,
  2022.

\bibitem[Peng et~al.(2025)Peng, Li, and Yang]{peng2025optimality}
Jingfu Peng, Yang Li, and Yuhong Yang.
\newblock On optimality of mallows model averaging.
\newblock \emph{Journal of the American Statistical Association}, 120\penalty0
  (550):\penalty0 1152--1163, 2025.

\bibitem[Polson and Scott(2012)]{polson2012local}
Nicholas~G. Polson and James~G. Scott.
\newblock Local shrinkage rules, {L\'evy} processes and regularized regression.
\newblock \emph{Journal of the Royal Statistical Society: Series B (Statistical
  Methodology)}, 74\penalty0 (2):\penalty0 287--311, 2012.
\newblock \doi{10.1111/j.1467-9868.2011.01015.x}.

\bibitem[Shao(1997)]{shao1997asymptotic}
Jun Shao.
\newblock An asymptotic theory for linear model selection.
\newblock \emph{Statistica sinica}, pages 221--242, 1997.

\bibitem[Shibata(1983)]{shibata1983asymptotic}
Ritei Shibata.
\newblock Asymptotic mean efficiency of a selection of regression variables.
\newblock \emph{Annals of the Institute of Statistical Mathematics},
  35\penalty0 (3):\penalty0 415--423, 1983.

\bibitem[Subramanian et~al.(2016)Subramanian, Ramsundar, Pande, and
  Denny]{subramanian2016bace}
Govindan Subramanian, Bharath Ramsundar, Vijay Pande, and Rajiah~Aldrin Denny.
\newblock Computational modeling of {$\beta$}-secretase 1 ({BACE-1}) inhibitors
  using ligand based approaches.
\newblock \emph{Journal of Chemical Information and Modeling}, 2016.
\newblock URL \url{https://pubmed.ncbi.nlm.nih.gov/27689393/}.

\bibitem[Thind et~al.(2020)Thind, Multani, and Cao]{thind2020functional}
Barinder Thind, Kevin Multani, and Jiguo Cao.
\newblock Neural networks as functional classifiers.
\newblock arXiv:2010.04305, 2020.
\newblock URL \url{https://github.com/caojiguo/FunClassifiers}.

\bibitem[Tibshirani(1996)]{tibshirani1996}
R~Tibshirani.
\newblock Regression shrinkage and selection via the {Lasso}.
\newblock \emph{Journal of the Royal Statistical Society: Series B
  (Methodological)}, 58\penalty0 (1):\penalty0 267--288, 1996.

\bibitem[Wang et~al.(2014)Wang, Paterlini, Gao, and Yang]{wang2014adaptive}
Zhan Wang, Sandra Paterlini, Fuchang Gao, and Yuhong Yang.
\newblock Adaptive minimax regression estimation over sparse {$\ell_q$}-hulls.
\newblock \emph{Journal of Machine Learning Research}, 15\penalty0
  (50):\penalty0 1675--1711, 2014.
\newblock URL \url{https://jmlr.org/papers/v15/wang14b.html}.

\bibitem[Xie(2017)]{xie2017heteroscedasticity}
Tian Xie.
\newblock Heteroscedasticity-robust model screening: A useful toolkit for model
  averaging in big data analytics.
\newblock \emph{Economics Letters}, 151:\penalty0 119--122, 2017.
\newblock \doi{10.1016/j.econlet.2016.12.019}.

\bibitem[Xu and Zhang(2025)]{xu2025model}
Wenchao Xu and Xinyu Zhang.
\newblock From model selection to model averaging: A comparison for nested
  linear models.
\newblock \emph{Econometric Theory}, pages 1--33, 2025.

\bibitem[Yang(1999)]{yang1999model}
Yuhong Yang.
\newblock Model selection for nonparametric regression.
\newblock \emph{Statistica Sinica}, pages 475--499, 1999.

\bibitem[Yang(2001)]{yang2001adaptive}
Yuhong Yang.
\newblock Adaptive regression by mixing.
\newblock \emph{Journal of the American Statistical Association}, 96\penalty0
  (454):\penalty0 574--588, 2001.

\bibitem[Yang(2007)]{yang2007consistency}
Yuhong Yang.
\newblock Consistency of cross validation for comparing regression procedures.
\newblock \emph{The Annals of Statistics}, 35\penalty0 (6):\penalty0
  2450--2473, 2007.
\newblock \doi{10.1214/009053607000000514}.

\bibitem[Yu et~al.(2025)Yu, Zhang, and Liang]{yu2025unified}
Dalei Yu, Xinyu Zhang, and Hua Liang.
\newblock Unified optimal model averaging with a general loss function based on
  cross-validation.
\newblock \emph{Journal of the American Statistical Association}, 120\penalty0
  (552):\penalty0 2697--2708, 2025.
\newblock \doi{10.1080/01621459.2025.2487215}.

\bibitem[Yuan and Yang(2005)]{yuan2005combining}
Zheng Yuan and Yuhong Yang.
\newblock Combining linear regression models: When and how?
\newblock \emph{Journal of the American Statistical Association}, 100\penalty0
  (472):\penalty0 1202--1214, 2005.

\bibitem[Zhang(2010)]{zhang2010minimax}
Cun-Hui Zhang.
\newblock Nearly unbiased variable selection under minimax concave penalty.
\newblock \emph{The Annals of Statistics}, 38\penalty0 (2):\penalty0 894--942,
  2010.
\newblock \doi{10.1214/09-AOS729}.

\bibitem[Zhang(2021)]{zhang2021new}
Xinyu Zhang.
\newblock A new study on asymptotic optimality of least squares model
  averaging.
\newblock \emph{Econometric Theory}, 37\penalty0 (2):\penalty0 388--407, 2021.

\bibitem[Zhang and Liu(2023)]{zhang2023prediction}
Xinyu Zhang and Chu-An Liu.
\newblock Model averaging prediction by {K}-fold cross-validation.
\newblock \emph{Journal of Econometrics}, 235\penalty0 (1):\penalty0 280--301,
  2023.
\newblock \doi{10.1016/j.jeconom.2022.04.007}.

\bibitem[Zhang et~al.(2016)Zhang, Yu, Zou, and Liang]{zhang2016optimal}
Xinyu Zhang, Dalei Yu, Guohua Zou, and Hua Liang.
\newblock Optimal model averaging estimation for generalized linear models and
  generalized linear mixed-effects models.
\newblock \emph{Journal of the American Statistical Association}, 111\penalty0
  (516):\penalty0 1775--1790, 2016.

\bibitem[Zhang et~al.(2020)Zhang, Zou, Liang, and
  Carroll]{zhang2020parsimonious}
Xinyu Zhang, Guohua Zou, Hua Liang, and Raymond~J Carroll.
\newblock Parsimonious model averaging with a diverging number of parameters.
\newblock \emph{Journal of the American Statistical Association}, 115\penalty0
  (530):\penalty0 972--984, 2020.

\bibitem[Zhu et~al.(2023)Zhu, Wang, Zhang, and Liang]{zhu2023scalable}
Rong Zhu, Haiying Wang, Xinyu Zhang, and Hua Liang.
\newblock A scalable frequentist model averaging method.
\newblock \emph{Journal of Business \& Economic Statistics}, 41\penalty0
  (4):\penalty0 1228--1237, 2023.
\newblock \doi{10.1080/07350015.2022.2116442}.

\end{thebibliography}
